%% file: root.tex
\documentclass[letterpaper, 10 pt, conference]{ieeeconf}  % Comment this line out if you need a4paper

\let\labelindent\relax

\usepackage[caption=false,font=scriptsize]{subfig}
\usepackage{tikz}
\usepackage[subpreambles=true]{standalone}
\usepackage{times}
\usepackage{epsfig}
\usepackage{graphicx}
\usepackage{amsmath}
\usepackage{amssymb}
\usepackage{booktabs}
\usepackage[separate-uncertainty=true,multi-part-units=single]{siunitx}
\usepackage{mathtools}
\usepackage{array}
\usepackage{multirow}
\usepackage{import}
\usepackage{placeins}
\usepackage[inline]{enumitem}
\usepackage{threeparttable}
\usepackage{pgfplots}
\makeatletter
\let\NAT@parse\undefined
\makeatother
\usepackage[hidelinks]{hyperref}

\pgfplotsset{compat=1.17}

\usetikzlibrary{calc}
\usetikzlibrary{positioning}
\usepgfplotslibrary{groupplots}

\definecolor{darkorange25512714}{RGB}{255,127,14}
\definecolor{lightgray204}{RGB}{204,204,204}
\definecolor{limegreen}{RGB}{50,205,50}
\definecolor{magenta}{RGB}{255,0,255}
\definecolor{steelblue31119180}{RGB}{31,119,180}

\IEEEoverridecommandlockouts                              % This command is only needed if 
\newcolumntype{L}[1]{>{\raggedright\let\newline\\\arraybackslash\hspace{0pt}}m{#1}}
\newcolumntype{C}[1]{>{\centering\let\newline\\\arraybackslash\hspace{0pt}}m{#1}}
\newcolumntype{R}[1]{>{\raggedleft\let\newline\\\arraybackslash\hspace{0pt}}m{#1}}

\DeclarePairedDelimiterX{\infdivx}[2]{(}{)}{%
  #1\;\delimsize\|\;#2%
}
\newcommand{\kldiv}{D_{\text{KL}}\infdivx}

\title{\LARGE \bf
A Probabilistic Framework for Visual Localization in Ambiguous Scenes
}

\author{Fereidoon Zangeneh$^{1,2}$, Leonard Bruns$^{1}$, Amit Dekel$^{2}$, Alessandro Pieropan$^{2}$ and Patric Jensfelt$^{1}$% <-this % stops a space
\thanks{* This work was partially supported by the Wallenberg AI, Autonomous Systems and Software Program (WASP) funded by the Knut and Alice Wallenberg Foundation. Authors thank Thien-Minh Nguyen for his help in recording and obtaining ground-truth poses for the new image sequence.}% <-this % stops a space
\thanks{$^{1}$ Authors are with the division of Robotics, Perception and Learning, KTH Royal Institute of Technology, SE-100\,44 Stockholm, Sweden.
        {\tt\small \{fzk,leonardb,patric\}@kth.se}}%
\thanks{$^{2}$ Authors are with Univrses AB, SE-118\,26 Stockholm, Sweden.
        {\tt\small \{firstname.lastname\}@univrses.com}}%
}

\begin{document}

\maketitle
\thispagestyle{empty}
\pagestyle{empty}

%%%%%%%%%%%%%%%%%%%%%%%%%%%%%%%%%%%%%%%%%%%%%%%%%%%%%%%%%%%%%%%%%%%%%%%%%%%%%%%%
\begin{abstract}
Visual localization allows autonomous robots to relocalize when losing track of their pose by matching their current observation with past ones. However, ambiguous scenes pose a challenge for such systems, as repetitive structures can be viewed from many distinct, equally likely camera poses, which means it is not sufficient to produce a single best pose hypothesis. In this work, we propose a probabilistic framework that for a given image predicts the arbitrarily shaped posterior distribution of its camera pose. We do this via a novel formulation of camera pose regression using variational inference, which allows sampling from the predicted distribution. Our method outperforms existing methods on localization in ambiguous scenes. Code and data will be released at  \href{https://github.com/efreidun/vapor}{github.com/efreidun/vapor}.

\end{abstract}

%%%%%%%%%%%%%%%%%%%%%%%%%%%%%%%%%%%%%%%%%%%%%%%%%%%%%%%%%%%%%%%%%%%%%%%%%%%%%%%%
\section{Introduction}

Visual localization is the task of inferring the ego pose of a camera from its image. It enables mobile robots to localize themselves in an environment, which is crucial for their navigation. Regardless of the paradigm that is followed to solve this task, the proposed methods revolve around detection of visual features that are unique to different regions of the environment and the camera poses that view them. Some methods do this by retrieving the most similar image to a query image from a database of images previously collected in the scene \cite{torii201524,arandjelovic2016netvlad,chen2017deep,hausler2021patch}; some establish point correspondences between the salient features of the query image and a pre-built 3D feature map, and use projective geometry relations to estimate the camera pose \cite{li2012worldwide,sattler2016efficient,liu2017efficient,sarlin2021back,sattler2012improving}; and some delegate this estimation problem to end-to-end learning-based solutions that regress the camera pose from what it views \cite{kendall2015posenet,clark2017vidloc,walch2017image,brahmbhatt2018geometry,chen2021direct}.

As long as there are unique identifying features in the images, there exist numerous solutions that can accurately estimate the camera pose \cite{sattler2016efficient, moreau2022lens}. However, the same cannot be said when the scene is ambiguous \cite{deng2022deep}, that is, when it contains distinct regions that are visually indistinguishable. Examples of this include identical doors, identical chairs arranged around a table, or the flights of stairs in a staircase, as illustrated in Fig.~\ref{fig:overview}. A desired solution in these cases is one that produces multiple pose hypotheses, capturing the repetitive patterns of the scene, rather than attempting to produce a single best hypothesis. This calls for a multi-hypothesis localization framework, which we address in this work. We focus on inference of the camera pose distribution from a single image, and refer to the rich literature on robot localization for how to accumulate evidence and maintain such a distribution over time \cite{fox2001particle,jensfelt2001active,Fox_NIPS2001_amcl}.

\definecolor{ferryblue}{rgb}{0.0, 0.4, 1.0}

\begin{figure}[t]
    \centering

    \scriptsize
    
    \begin{tikzpicture}
        
        \node[inner sep=0.5\pgflinewidth, outer sep=0pt, draw=ferryblue, ultra thick, anchor=west] (q1) at (0,0) {\includegraphics[width=1.5cm]{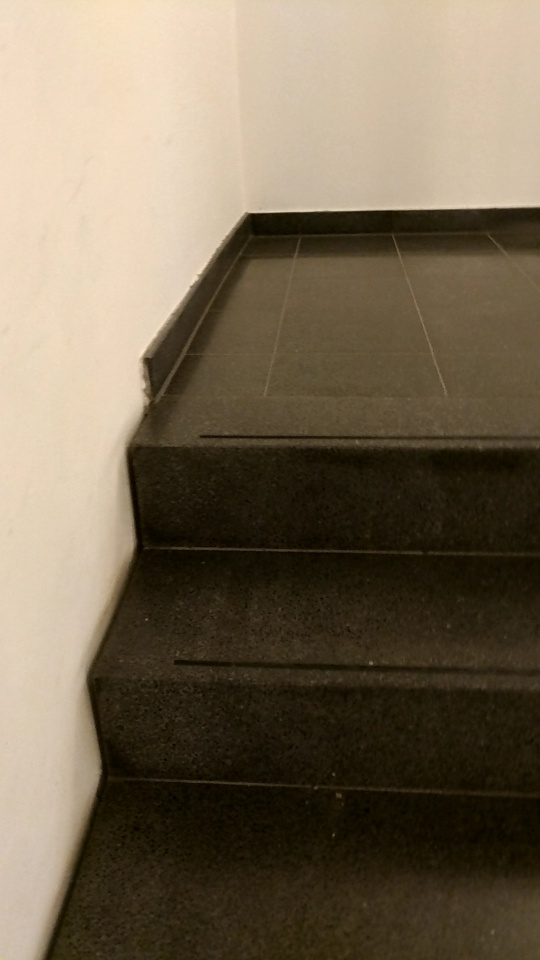}};
        \node[inner sep=0.5\pgflinewidth, outer sep=0pt, draw=green, ultra thick,anchor=west] (q2) at ($(q1.east) + (0.2,0)$) {\includegraphics[width=1.5cm]{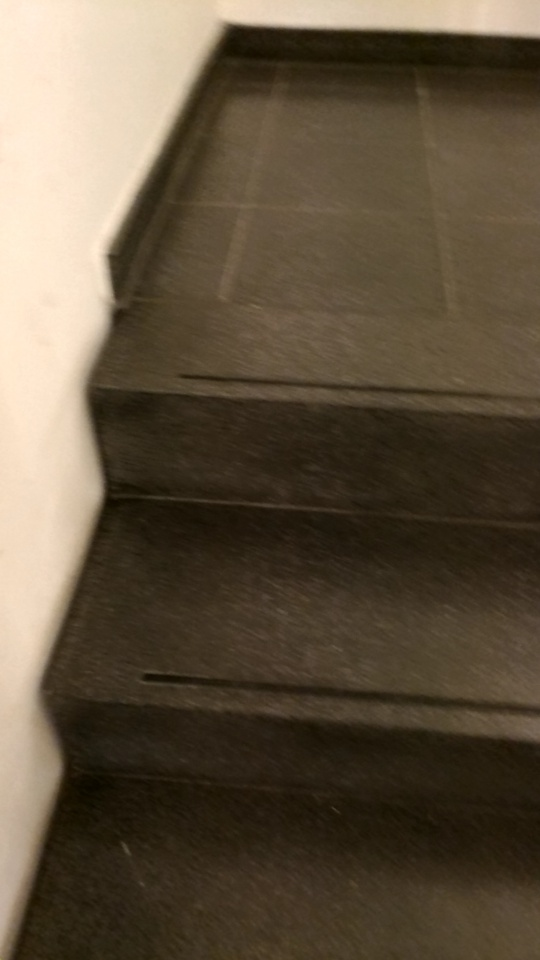}};
        \node[inner sep=0.5\pgflinewidth, outer sep=0pt, draw=red, ultra thick,anchor=west] (q3) at ($(q2.east) + (0.2,0)$) {\includegraphics[width=1.5cm]{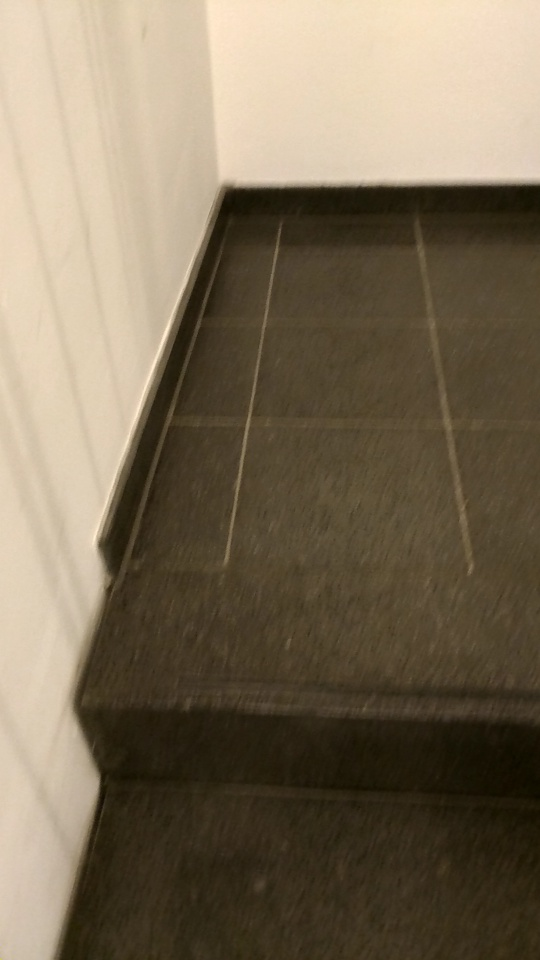}};

        \node[draw, anchor=west, align=center, very thick] (vapor) at ($(q3.east) + (0.3,0.0)$) {Variational\\Pose Regressor};
        
        \draw[very thick, ->] (q3.east) -- (vapor);
        \node[inner sep=0.5\pgflinewidth, outer sep=0pt, draw=red, ultra thick,anchor=west] (q3) at ($(q2.east) + (0.2,0)$) {\includegraphics[width=1.5cm]{figures/staircase_example/query_2.png}};
        
        \node[inner sep=0pt, outer sep=2pt, anchor=west, rounded corners=0.5cm] (position) at (0, -3.7) {\includegraphics[width=3cm, angle=180]{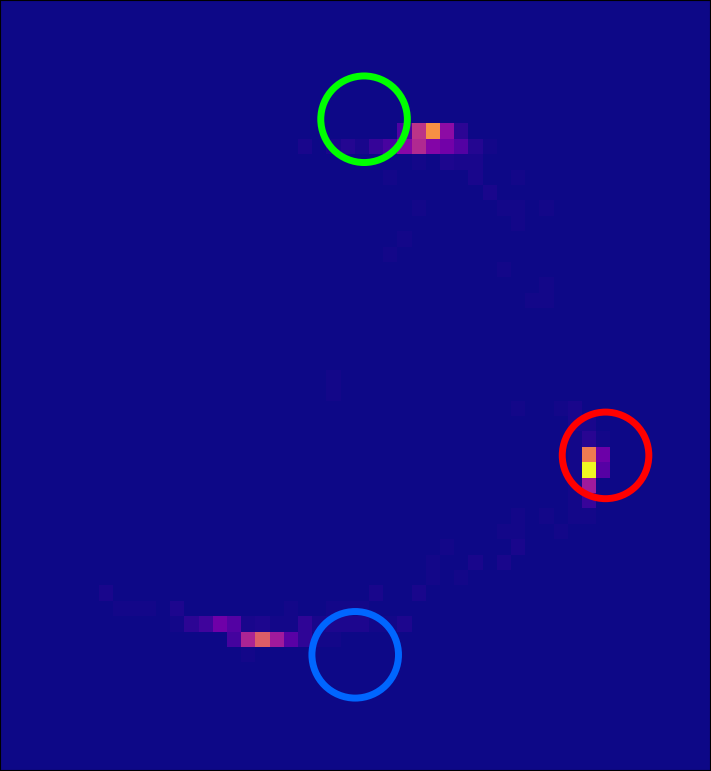}};
        \node[inner sep=0pt, outer sep=2pt, anchor=north] (orientation) at (position.south) {\includegraphics[width=3cm]{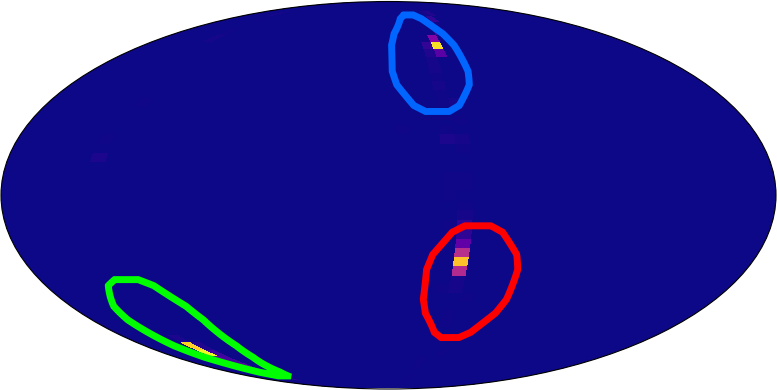}};
        \node[inner sep=0pt, outer sep=2pt, anchor=north west] (3d) at (position.north east) {\includegraphics[width=4.7cm]{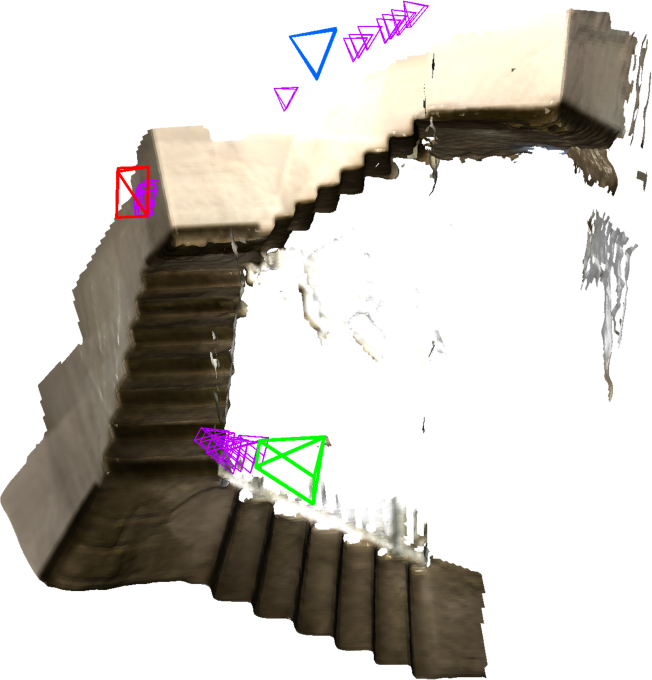}};

        \coordinate (distributionbox) at ($(position.north west) + (-0.35, 0.1)$);
        \draw[very thick, rounded corners=0.3cm] (distributionbox) rectangle ($(3d.south east) + (0.15, -0.45)$);

        \draw[very thick, ->] (vapor.east) to[out=0, in=90, looseness=1] ++(0.6, -0.6) to [out=-90, in=0, looseness=1] ++(-0.6, -0.6) to[out=180,in=90, looseness=1] ($(distributionbox) + (7,0)$);

        \node[below, align=center,outer sep=2pt] (imagelabel) at (q2.south) {(a) Ambiguous images};
        \node[below, align=center] (imagelabel) at (orientation.south) {(b) Pose posterior};
        \node[below, align=center] (imagelabel) at (3d.south) {(c) Samples from pose posterior};

        \node[inner sep=0pt, outer sep=0pt, anchor=east] at (position.west) {\rotatebox{90}{Position}};
        
        \node[inner sep=0pt, outer sep=0pt, anchor=east] at (orientation.west) {\rotatebox{90}{Orientation}};

        \draw[white, thick, ->] ($(position.south west) + (0.1,0.2)$) -- ($(position.south east) + (-0.33,0.2)$) node[right, outer sep=1pt, inner sep=0pt] {$x$};
        \draw[white, thick, ->] ($(position.south west) + (0.2,0.1)$) -- ($(position.north west) + (0.2,-0.36)$) node[above, outer sep=1pt, inner sep=0pt] {$y$};
        
    \end{tikzpicture}
    \caption{(a) Visually similar images taken from three different flights of stairs, (b) camera pose distribution predicted for the right image, and (c) samples drawn from this posterior. The distribution is visualized by a position heatmap on the $xy$-plane (marginalizing height) and an elliptic orientation heatmap\protect\footnotemark. We show the drawn samples in a 3D reconstruction of the scene by small camera frusta in purple. The ground-truth camera poses are shown by color-coded circles in (b) and camera frusta in (c).}\label{fig:overview}
\end{figure}

\footnotetext{We use the Mollweide projection for the surface of the 2-sphere component of $\mathrm{SO}(3)$ obtained through Hopf-fibration (marginalizing the fibers), inspired by Murphy et al.\ \cite{murphy2021implicit}.}

We propose a probabilistic framework that allows inferring the posterior distribution over camera poses for a given image. We represent this distribution by an arbitrary number of samples drawn from it, which in theory can model distributions with any number of modes and of any shape. Samples from this distribution can be used in downstream tasks, such as motion planning or active localization. We formulate our solution following the paradigm of end-to-end camera pose regression, and employ variational inference \cite{kingma2014auto,rezende2014stochastic} to model the visual features of images used for localization. We show that camera pose regression, despite its limitations in generalization and accuracy compared to structure-based methods \cite{sattler2019understanding}, when combined with variational inference gives rise to a simple, yet powerful solution for pose posterior prediction from an observed image.

We summarize our contributions as the following:
\begin{enumerate*} [label=(\arabic*)]
    \item We lay out a novel formulation of camera pose regression using variational inference, which allows sampling from an arbitrarily shaped pose distribution for a given image.
    \item We propose a novel sampling-based Winners-Take-All optimization scheme, which allows learning multimodal distributions.
    \item We record a sequence of real-world camera images capturing a case of severe visual ambiguity for evaluation of localization solutions.
    \item We show that our formulation outperforms existing methods on ambiguous scenes.
\end{enumerate*}

%-------------------------------------------------------------------------
\section{Related work}
\label{sec:related}

Regression-based approaches aim to solve the pose estimation problem in a single step by finding a function that directly maps an image to its pose, promising improved performance in feature-less environments or under motion blur \cite{walch2017image}. In early work, Shotton et al.\  \cite{shotton2013scene} proposed regressing 3D scene coordinates for each pixel in an image. In combination with depth data, this allows robust estimation of the 6D pose of the camera by employing RANSAC with Kabsch's algorithm \cite{kabsch1976solution}. The first end-to-end approach for image-based pose regression was PoseNet proposed by Kendall et al.\  \cite{kendall2015posenet}. Specifically, they proposed to train a deep neural network to directly regress the 6D camera pose from the image features extracted by a pre-trained backbone.

Following this early work, various improvements orthogonal to our work have subsequently been proposed. Naseer and Burgard \cite{naseer2017deep} showed that RGB-D data can be exploited to generate additional views from the limited training images to improve performance. Recently, Ng et al.\  \cite{ng2021reassessing} and Moreau et al.\  \cite{moreau2022lens} extended this idea to RGB data. Other works propose to use additional information often available in robotic applications \cite{clark2017vidloc,brahmbhatt2018geometry}.

More closely related to our work, several works investigate how to model uncertainty for pose regressors. In \cite{kendall2016modelling}, the authors apply Bayesian deep learning to PoseNet. This allows one to gauge the uncertainty in the prediction, although the ability to learn more complicated distributions remains limited, as noted by \cite{deng2022deep}. While \cite{kendall2016modelling} focused on epistemic uncertainty, Kendall and Cipolla\  \cite{kendall2017geometric} considered homoscedastic aleatoric uncertainty by modifying the loss, and Moreau et al.\  \cite{moreau2022coordinet} modeled heteroscedastic aleatoric uncertainty instead by predicting an uncertainty measure.

% TODO check if huang2019 drop out for dynamic scenes is relevant

Deng et al.\  \cite{deng2022deep} further extend the idea of representing uncertainty by predicting a mixture of multiple unimodal distributions. In principle, this allows the network to correctly predict multiple modes for ambiguous queries. A downside to this mixture-based approach is the difficulty of picking the correct number of modes, and training the network so that it actually predicts different modes. To handle the latter issue, the authors propose a Winner-Takes-All scheme that only gives supervision to the best predicted mode. Our work follows a similar idea, but instead of employing a mixture model with a fixed number of components, we follow a variational approach, which, in principle, can learn to produce arbitrarily shaped pose distributions.

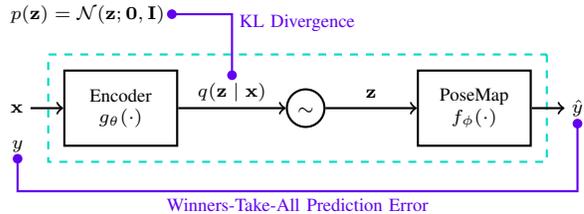
\begin{figure}[t]
    \centering
    \subimport{figures/}{pipeline.tex}
    % \vspace{-0.3cm}
    \caption{Our pipeline for inference of the camera pose distribution for an image $\mathbf{x} \in \mathbb{R}^{H \times W \times 3}$ with ground-truth pose label $y \in \mathrm{SE}(3)$. We can simulate the posterior distribution $p(y \mid \mathbf{x})$ by drawing samples $\mathbf{z} \sim q(\mathbf{z} \mid \mathbf{x}), \mathbf{z} \in \mathbb{R}^d$ and applying the mapping $f_\phi(\mathbf{z})$ to get $\hat{y} \sim p(y \mid \mathbf{x}), \hat{y} \in \mathrm{SE}(3)$. The loss terms used in the learning objective are shown in purple.}\label{fig:pipeline}
\end{figure}

In another related line of work, Murphy et al.\  \cite{murphy2021implicit} build on the recent success of neural fields \cite{xie2022neural} by employing an MLP that predicts the probability density for a given rotation, allowing representation of arbitrary distributions in $\mathrm{SO}(3)$. Our work also aims to learn arbitrary distributions, but we propose a sampling-based approach, in which a sample from a latent space is transformed to a pose in $\mathrm{SE}(3)$. This simplifies inference, as it does not require dense querying of the support to find the modes of the distribution; instead, our approach allows direct sampling from it.

\section{Method}

We propose to perform visual localization for an image in two steps:
\begin{enumerate*} [label=(\arabic*)]
    \item infer a distribution in the latent space capturing the visual features that are useful for localization within the scene;
    \item perform a random variable transformation to obtain a distribution of camera poses for the query image.
\end{enumerate*}
Fig.~\ref{fig:pipeline} visualizes our proposed pipeline.

\subsection{Formulation}

Let $\mathbf{x} \in \mathbb{R}^{H \times W \times 3}$ be a color image taken from camera pose $y \in \mathrm{SE}(3)$. In localization, where the scene is known beforehand, one can in theory infer the posterior distribution of visual features as seen in the observed image $p(\mathbf{z} \mid \mathbf{x})$. Here, $\mathbf{z} \in \mathbb{R}^d$ is the latent variable corresponding to the visual features that the scene comprises. With this definition of the latent variable, visually similar images result in similar posterior distributions in the latent space, even if the images are taken from distinct camera poses, as in ambiguous scenes.

Having full knowledge of the scene, the posterior distribution of visual features should contain the information needed to infer the posterior distribution of camera poses given the observed image $p(y \mid \mathbf{x})$. This can be formulated as a transformation of densities from visual features in $\mathbb{R}^d$ to camera pose in $\mathrm{SE}(3)$, which can be achieved by applying a deterministic mapping $f : \mathbb{R}^d \to \mathrm{SE}(3)$ to samples drawn from the posterior distribution in the latent space: $y = f(\mathbf{z}), ~~\mathbf{z} \sim p(\mathbf{z} \mid \mathbf{x})$.

\subsection{Modeling via learning}

In the proposed formulation, there are two scene-dependent operations that model the scene for the purpose of visual localization, namely the inference of the posterior distribution in the latent visual features' space $p(\mathbf{z} \mid \cdot)$, and the mapping to camera pose $f(\cdot)$. We parameterize these in the weights of two deep neural networks and learn them from data samples collected from the scene. We refer to the two networks as \textit{Encoder} $g_\theta(\cdot)$ and \textit{PoseMap} $f_\phi(\cdot)$, parameterized by $\theta$ and $\phi$, respectively.

\textbf{Encoder} $g_\theta(\cdot)$ is an inference network with a Gaussian inference model that for an input image $\mathbf{x}$ outputs $\boldsymbol{\mu} \in \mathbb{R}^d$ and $\boldsymbol{\sigma} \in \mathbb{R}^d$ defining the posterior distribution $q(\mathbf{z} \mid \mathbf{x}) = \mathcal{N}(\mathbf{z} ; \boldsymbol{\mu}, \text{diag}(\boldsymbol{\sigma}^2))$ in the latent space. This follows the variational principle, where an unknown posterior distribution is modeled by optimizing the parameters of a convenient family of distributions such as Gaussians to best resemble the true posterior. Akin to Variational Auto-Encoders (VAEs) \cite{kingma2014auto,rezende2014stochastic}, we amortize this per-image optimization at inference time by optimizing the \textit{Encoder} weights at training time to directly predict the distribution parameters.

\textbf{PoseMap} $f_\phi(\cdot)$ is a fully connected network that, for an input sample from the latent space $\mathbf{z}$, outputs a camera pose $y$. This means that the posterior distribution of the camera pose $p(y \mid \mathbf{x})$ can be approximated by simulating the inferred posterior distribution in the latent space $\mathbf{z} \sim q(\mathbf{z} \mid \mathbf{x})$ via reparameterization trick and passing the drawn samples through the mapping $y = f_\phi(\mathbf{z})$ to obtain samples $y \sim p(y \mid \mathbf{x})$. The output of the network $y$ comprises a translation vector $\mathbf{t} \in \mathbb{R}^3$ and a 6D representation for rotation $\mathbf{r} \in \mathbb{R}^6$. The rotation parameterization choice is the continuous representation for rotations in 3D introduced by Zhou et al.\  \cite{zhou2019continuity}, where a rotation matrix is retrieved from the 6D representation following a Gram-Schmidt-like process. 

\subsection{Learning scheme}

The network weights $\theta, \phi$ that represent a scene are learned from a dataset of images and camera poses $\mathcal{D} = \{(\mathbf{x}_i, y_i) \mid i=1, \ldots, N\}$ collected in that scene. For this, we lay out an optimization scheme that enables learning multimodal pose distributions as is desired in ambiguous scenes.

\subsubsection{Objective terms}

\textbf{Prediction error} measures the quality of a predicted pose $\hat{y} \in \mathrm{SE}(3)$ against its ground truth $y$. We define the prediction error as the weighted sum of a translation error term defined on $\mathbb{R}^3$ and a rotation error term defined on $\mathrm{SO}(3)$. The translation error is the Euclidean distance between the translation components $\hat{\mathbf{t}}, \mathbf{t} \in \mathbb{R}^3$ of predicted and ground-truth poses. For the rotation error, we opt for the chordal distance between the rotation components $\hat{\mathbf{R}}, \mathbf{R} \in \mathrm{SO}(3)$, for its more favorable gradients in gradient-based optimization than, for example, the geodesic distance's. The prediction error is thus defined as
\begin{equation}
    d_{\text{pose}}(\hat{y}, y) = \lambda_t \lVert \hat{\mathbf{t}} - \mathbf{t} \rVert_2 + \lambda_r \lVert \hat{\mathbf{R}} - \mathbf{R} \rVert_\mathrm{F},
\end{equation}
where $\lambda_t$ and $\lambda_r$ are tunable constants, balancing the scales of the two terms.

\textbf{Kullback–Leibler divergence} $\kldiv{q(\mathbf{z} \mid \mathbf{x})}{p(\mathbf{z})}$ measures how different an inferred latent posterior distribution $q(\mathbf{z} \mid \mathbf{x})$ is from a prior distribution defined on the latent variable $p(\mathbf{z})$. This is an integral part of the variational principle, which together with the prediction error forms the evidence lower bound (ELBO) optimized in variational approaches. As is common practice, we assume a standard Gaussian prior $p(\mathbf{z}) = \mathcal{N}(\mathbf{z} ; \mathbf{0}, \mathbf{I})$ for its simplicity in computing the KL divergence.

\subsubsection{Evidence lower bound (ELBO)} \label{sec:problem_with_elbo}

In variational approaches, the ELBO objective that is typically maximized is a combination of negative KL divergence and expected log-likelihood of predictions $\mathbb{E}_{q(\mathbf{z} \mid \mathbf{x})}[\log p_\phi(y \mid \mathbf{z})]$. The latter expectation is generally computed by Monte Carlo simulation of $q(\mathbf{z} \mid \mathbf{x})$. With our choice of pose prediction error, the variational optimization objective can be written as
\begin{equation}
    \begin{split}
        \min_{\theta, \phi} \sum_{\mathbf{x}_i, y_i \in \mathcal{D}} \Bigg[&\kldiv{q(\mathbf{z} \mid \mathbf{x}_i)}{p(\mathbf{z})} \\& + \frac{1}{|\mathcal{Z}_i|}\sum_{\mathbf{z}_j \in \mathcal{Z}_i}d_{\text{pose}}(f_\phi(\mathbf{z}_j), y_i)\Bigg],
    \end{split}
\end{equation}
where $\mathcal{Z}_i =\{\mathbf{z}_j \sim q(\mathbf{z} \mid \mathbf{x}_i) \mid j=1, \ldots, M\}$ is the Monte Carlo sample set and $|\mathcal{Z}_i|$ its cardinality.

We argue that minimizing this objective, and specifically the expected prediction error, is counterproductive in our setting, where the camera pose posterior $p(y \mid \mathbf{z})$ can be multimodal in ambiguous scenes. In such scenarios, two visually similar images $\mathbf{x}_i$ and $\mathbf{x}_j$ $(i \neq j)$ are encoded to similar latent posterior distributions $p(\mathbf{z} \mid \mathbf{x}_i)$ and $p(\mathbf{z} \mid \mathbf{x}_j)$. However, these images can be taken from two distinct poses $y_i$ and $y_j$ in the scene, in which case the true posterior distributions of the camera pose $p(y \mid \mathbf{x}_i)$ and $p(y \mid \mathbf{x}_j)$ are both bimodal. Minimizing the expected prediction error results in a compromised solution in the form of a unimodal inferred distribution between the two true modes. We propose a modification of the expected error term to address this.

\subsubsection{Winners-Take-All optimization}

We propose to confine the computed mean prediction error to a subset of Monte Carlo samples $\hat{\mathcal{Z}}_i \subseteq \mathcal{Z}_i$, whose image through the mapping $f_\phi(\cdot)$ is within a certain distance $\delta$ of the true mode $y_i$, that is, $\hat{\mathcal{Z}}_i = \{\mathbf{z}_j \in \mathcal{Z}_i \mid d_{\text{pose}}(f_\phi(\mathbf{z}_j), y_i) < \delta\}$. This ensures that pose samples can concentrate around individual modes during optimization without influence from other modes. However, the true posterior is unknown and different modes can have different shapes, rendering the choice of $\delta$ non-trivial. Moreover, random initialization of the parameters $\theta$ and $\phi$ does not guarantee that there will be pose samples within any $\delta$ distance of the modes at the start of the optimization. This calls for an adaptive selection of $\delta$ at every iteration and for every mode.

At every iteration and for a ground-truth pose $y_i$ we pick $\delta_{i,\alpha}$ as the radius of the smallest ball centered at $y_i$ containing a fraction $\alpha$ of samples in $\mathcal{Z}_i$. In other words, our adaptive $\delta_{i,\alpha}$, defined as
\begin{equation}
    \begin{split}
        \delta_{i,\alpha} = \inf \bigg\{ \delta \in \mathbb{R}_+ \Bigm\vert \big|\hat{\mathcal{Z}}_i\big| = \left\lfloor\alpha \cdot|\mathcal{Z}_i|\right\rfloor \bigg\},
    \end{split}
\end{equation}
results in minimizing the prediction error for only the closest fraction $\alpha$ of Monte Carlo samples per ground-truth pose $y_i$. Our proposed optimization objective is
\begin{equation}
    \begin{split}
        \min_{\theta, \phi} \sum_{\mathbf{x}_i, y_i \in \mathcal{D}} \Bigg[&\beta ~\kldiv{q(\mathbf{z} \mid \mathbf{x}_i)}{p(\mathbf{z})} \\& + \frac{1}{|\hat{\mathcal{Z}}_{i, \alpha}|}\sum_{\mathbf{z}_j \in \hat{\mathcal{Z}}_{i, \alpha}}d_{\text{pose}}(f_\phi(\mathbf{z}_j), y_i)\Bigg],
    \end{split}
\end{equation}
where $\hat{\mathcal{Z}}_{i, \alpha} = \{\mathbf{z}_j \in \mathcal{Z}_i \mid d_{\text{pose}}(f_\phi(\mathbf{z}_j), y_i) < \delta_{i,\alpha}\}$. $\alpha$ and $\beta$ are tunable constants, the latter being the balancing weight for the KL divergence term.

This is in spirit similar to the Winner-Takes-All multi-hypothesis optimization scheme used for learning mixture models, where the closest mixture component is optimized per label \cite{deng2022deep, makansi2019overcoming}. However, our proposed solution is in a different setting, as we represent posteriors by samples instead of mixture models. We therefore refer to our method as Winners-Take-All to acknowledge this similarity, while reflecting the fact that it is used for optimizing sample sets rather than individual mixture components.

\section{Experiments}
\label{sec:result}

\subsection{Implementation details}

We implement our method using the PyTorch library \cite{paszke2019pytorch}. We use ResNet-18 \cite{he2016deep} as the backbone of the \textit{Encoder} to extract $2048$-dimensional feature vectors, followed by a linear layer to predict $d$-dimensional $\boldsymbol{\mu}$ and $\log\boldsymbol{\sigma}^2$ vectors for the latent posterior. The \textit{PoseMap} is implemented with a fully connected network taking the input vector through the dimensionality transformation $d \rightarrow 128 (\rightarrow 128)_{\times n_\text{layers}} \rightarrow 3 + 6$ with ReLU activations in-between. The minimum number of hidden layers $n_\text{layers}$ depends on the complexity of the target pose distributions in the scene. In nearly all tested scenes we achieved favorable performance with as few as $n_\text{layers} = 3$, which, unless otherwise stated, is used across all experiments. The final layer corresponds to the prediction of translation and rotation vectors, where the former goes through a sigmoid activation, followed by a fixed affine transformation that shifts and scales the predictions to the metric ranges of the scene.

We train our networks using Adam optimizer \cite{kingma2014adam} with initial learning rate of $\num{1e-4}$ and an exponential learning rate decay of $0.8$, applied every $n_\text{lr-decay}$ epochs for 10 occurences. Following the pose regression literature, we first resize each image such that its smallest edge is $256$, then randomly crop $224\times224$ regions for input to the \textit{Encoder}. We also augment the data with color\slash brightness jittering and Gaussian blur to account for lighting changes and motion blur between images. Unless otherwise stated, we let $\alpha=0.20$, $\beta=0.01$, use a $d=16$-dimensional latent space, and represent distributions with $1000$ Monte Carlo samples in all experiments, since we found this to produce good predictions in our setting. Other hyperparameters are reported in Table \ref{tab:implementation}, tuned to reflect the number of images and metric scales of different datasets, which range from small indoor to large outdoor scenes. Note that we found these settings without a major hyperparameter search, and one may improve the performance by a thorough search of the optimal hyperparameters.

\begin{table}[t]
    \centering
    \caption{Hyperparameters used in training}\label{tab:implementation}
    \scriptsize
    \begin{tabular}{@{}lccccc@{}}
        \toprule
        Dataset & $\lambda_t$ & $\lambda_r$ & Batch Size & $\#$ Epochs & $n_\text{lr-decay}$\\
        \midrule
        7-Scenes \cite{shotton2013scene} & $5$ & $10$ & $64$ & $100$ & $10$\\
        Cambridge Land. \cite{kendall2015posenet} & $5$ & $100$ & $64$ & $500$ & $50$\\
        Ambiguous Reloc. \cite{deng2022deep} & $5$ & $2$ & $4$ & $500$ & $50$\\
        Ceiling & $5$ & $2$ & $4$ & $2000$ & $50$\\
        Synthetic & $5$ & $2$ & $4$ & $500$ & $50$\\
        \bottomrule
    \end{tabular}
\end{table}

\subsection{Datasets and metrics}

We evaluate our method on the Ambiguous Relocalization dataset \cite{deng2022deep} as an existing benchmark with real-world image sequences of ambiguous environments. For each scene in the dataset there are separate training and test image sequences recorded from their own unique camera trajectories, but with generally similar views. We found that despite the apparent ambiguity to the human eye, a large fraction of frames in this dataset contain unique identifying features, which an expressive feature detector can infer the pose from. This results in unimodal predicted posteriors for a large number of frames, which hinders the evaluation of a method's capability in forming multimodal distributions. To address this, we complement the dataset by recording a new real-world sequence of a ceiling with machine-fabricated panels, capturing a case of severe visual ambiguity. We record the training and test sequences with a calibrated LiDAR-IMU-camera rig, and obtain ground-truth camera poses using MILIOM \cite{nguyen2021miliom}. We also render image sequences of two synthetic scenes from 3D Warehouse\footnote{https://3dwarehouse.sketchup.com/}, which contain symmetries by design, and use them to investigate our method in a controlled setting.

We use recall as the metric to evaluate pose distributions in ambiguous scenes. For a query image, we draw samples from its posterior distribution, and consider it a true positive if at least a fraction $\gamma$ of the samples are within a distance of the ground-truth pose (and a false negative otherwise). We argue that for a distribution with well-separated equally likely modes, setting $\gamma$ inversely proportional to the number of modes gives an estimate of whether the distribution contains sufficient density around the ground-truth pose. We report recall with $\gamma=0.1$ for all tested scenes except for the ceiling scene, where we use $\gamma = 0.05$.

To validate the performance of our method as a general pose regressor on unambiguous scenes, we evaluate it on the visual localization benchmarks 7-Scenes \cite{shotton2013scene} and Cambridge Landmarks \cite{kendall2015posenet}. As is commonly reported by pose regression works, we use median error for evaluation on these datasets. We obtain a point prediction from the Monte Carlo samples of each predicted distribution using the arithmetic and chordal $L_2$ \cite{hartley2013rotation} means for translation and orientation, respectively. The median of this estimate's error compared to the ground-truth pose is reported across each scene.

\begin{figure*}[t]
    \centering%
    \subimport{figures/ceiling}{007450.tex}%
    \vspace{-0.3cm}
    \caption[test]{Marginal posterior distributions along $x$-axis (top left) and $y$-axis (bottom right) predicted by our method (\tikz[baseline=0,steelblue31119180,opacity=0.9]{\draw[fill,draw=none] (0,0) rectangle (0.099,0.05) (0.1,0) rectangle (0.199,0.2) (0.2,0) rectangle (0.3,0.12);}), by Bingham MDN\ \cite{deng2022deep} (\tikz[baseline=0,darkorange25512714,opacity=0.7]{\draw[fill,draw=none] (0,0) rectangle (0.099,0.05) (0.1,0) rectangle (0.199,0.2) (0.2,0) rectangle (0.3,0.12);}), the prediction by MapMet\  \cite{brahmbhatt2018geometry} (\tikz [baseline=-0.5ex]{\draw[magenta,line width=1pt] (0,0) -- ++ (0.4,0);}), and the ground truth (\tikz [baseline=-0.5ex]{\draw[limegreen,line width=1pt] (0,0) -- ++ (0.4,0);}) for a query image (top right) from the ceiling scene. The heatmap shows the 2D histogram predicted by our method overlaid on top of stitched images of the scene. Note that our method successfully captures all six modes of the distribution while MapNet only predicts a single estimate at a wrong location, and Bingham MDN method assigns large probabilities in visually dissimilar locations.}\label{fig:posteriors}%
\end{figure*}

\subsection{Evaluation on benchmark datasets}

We report the results on the ambiguous scenes in Table \ref{tab:ambiguous}. We can see that our method, outperforms Bingham MDN \cite{deng2022deep} as the method closest to ours that predicts a distribution of poses aimed at localization in ambiguous scenes. We considered two settings of their approach with 10 and 50 components in their mixture model, and evaluated the metric based on samples drawn from them. As the 10-component setting consistently performed better, we report its results as a representative in the table (marked BMDN). Fig.~\ref{fig:posteriors} shows an example of the predicted posterior given a query image from the ceiling scene, where we can see posterior predicted by our method better captures the ambiguous structure of the scene. We also evaluate PoseNet \cite{kendall2015posenet} and its Bayesian variant \cite{kendall2016modelling}, as well as MapNet \cite{brahmbhatt2018geometry}. However, we see that these single estimate methods fail to achieve comparable performance on the ambiguous scenes. To our surprise, vanilla PoseNet performed comparatively better than Bayesian PoseNet, so we include its results as representative (marked PN) alongside MapNet (marked MN).

\begin{table}[t]
    \centering
    \begin{threeparttable}
    \caption{Measured recall in ambiguous scenes (higher is better)}\label{tab:ambiguous}
    \scriptsize
    %\begin{tabular}{@{}lrC{1.2cm}C{1.2cm}cC{1.2cm}C{1.2cm}cC{1.2cm}C{1.2cm}@{}}%
    \setlength{\tabcolsep}{4.0pt}
    \begin{tabular}{@{}lcccccc@{}}
        \toprule
        Scene & Threshold & PN \cite{kendall2015posenet} & MN \cite{brahmbhatt2018geometry} & BMDN \cite{deng2022deep} & Abl. & Ours\\
        \midrule
        \multirow{3}{*}{Blue Chairs} & $0.1\text{m} / 10^{\circ}$ & $0.08$ & $0.05$ & $0.41$ & $0.32$ & $\mathbf{0.45}$ \\
        & $0.2\text{m} / 15^{\circ}$ & $0.40$ & $0.33$ & $0.83$ & $0.89$ & $\mathbf{0.99}$ \\
        & $0.3\text{m} / 20^{\circ}$ & $0.56$ & $0.46$ & $0.89$ &$0.97$ & $\mathbf{1.00}$ \\
        \midrule
        \multirow{3}{*}{Meeting Table} & $0.1\text{m} / 10^{\circ}$ & $0.00$ & $0.00$ & $\mathbf{0.09}$ & $0.03$ & $0.06$ \\
        & $0.2\text{m} / 15^{\circ}$ & $0.02$ & $0.03$ & $0.27$ & $0.24$ & $\mathbf{0.35}$ \\
        & $0.3\text{m} / 20^{\circ}$ & $0.02$ & $0.07$ & $0.33$ & $0.34$ & $\mathbf{0.43}$ \\
        \midrule
        \multirow{3}{*}{Staircase} & $0.1\text{m} / 10^{\circ}$ & $0.00$ & $0.07$ & $0.24$ & $0.12$ & $\mathbf{0.30}$ \\
        & $0.2\text{m} / 15^{\circ}$ & $0.01$ & $0.17$ & $0.48$ & $0.44$ & $\mathbf{0.62}$ \\
        & $0.3\text{m} / 20^{\circ}$ & $0.01$ & $0.29$ & $0.69$ & $0.63$ & $\mathbf{0.72}$ \\
        \midrule
        \multirow{3}{*}{Staircase Ext.} & $0.1\text{m} / 10^{\circ}$ & $0.00$ & $0.01$ & $0.11$ & $0.03$ & $\mathbf{0.12}$ \\
        & $0.2\text{m} / 15^{\circ}$ & $0.00$ & $0.03$ & $0.43$ & $0.24$ & $\mathbf{0.53}$ \\
        & $0.3\text{m} / 20^{\circ}$ & $0.01$ & $0.07$ & $0.60$ & $0.44$ & $\mathbf{0.71}$ \\
        \midrule
        \multirow{3}{*}{Seminar Room} & $0.1\text{m} / 10^{\circ}$ & $0.00$ & $0.09$ & $0.38$ & $0.17$ & $\mathbf{0.43}$ \\
        & $0.2\text{m} / 15^{\circ}$ & $0.02$ & $0.37$ & $0.79$ & $0.53$ & $\mathbf{0.90}$ \\
        & $0.3\text{m} / 20^{\circ}$ & $0.10$ & $0.53$ & $0.91$ & $0.80$ & $\mathbf{0.97}$ \\
        \midrule
        \multirow{3}{*}{Ceiling$\smash{^\dagger}$} & $0.1\text{m} / 10^{\circ}$ & $0.00$ & $0.02$ & $0.08$ & $0.00$ & $\mathbf{0.09}$ \\
        & $0.2\text{m} / 15^{\circ}$ & $0.03$ & $0.05$ & $0.19$ & $0.02$ & $\mathbf{0.31}$ \\
        & $0.3\text{m} / 20^{\circ}$ & $0.06$ & $0.09$ & $0.30$ & $0.04$ & $\mathbf{0.44}$ \\
        \bottomrule
    \end{tabular}
    \begin{tablenotes}
         \item[$\dagger$] We train the independently recorded ceiling scene with $\alpha=0.05$ and $n_\text{layers}=9$, reflecting the richer presence of ambiguities in the scene.
    \end{tablenotes}
    \end{threeparttable}
\end{table}

In order to investigate whether our method's improved performance stems from our novel formulation with variational inference, we perform an ablation, in which we modify our pipeline to produce a single pose for an input image. We remove the KL divergence term from the objective, modify the \textit{Encoder} to predict a single point, and obtain a single pose prediction by passing the encoder's prediction through \textit{PoseMap}. All else equal, we evaluate this ablative variant of our method that is in principle very similar to PoseNet. We can see in Table \ref{tab:ambiguous} that this variant, marked \textit{Abl.}, while performing better than PoseNet due to its more recent feature extractor network, falls short of the unablated variant, validating the merit of our proposed formulation.

\begin{table}[t]
    \centering
    \begin{threeparttable}
        % \caption{Median translation and orientation errors (m / $^{\circ}$) in unambiguous scenes.}\label{tab:median_error}
        \caption{Median error (m / $^{\circ}$) in unambiguous scenes (lower is better)}\label{tab:median_error}
        \scriptsize
        % \begin{tabular}{@{}clC{1.8cm}C{2cm}C{2.5cm}C{2.0cm}c@{}}
        \setlength{\tabcolsep}{4.0pt}
        \begin{tabular}{@{}lccccc@{}}
            \toprule
            Scene & PN \cite{kendall2015posenet} & MN$\smash{^\dagger}$ \cite{brahmbhatt2018geometry} & BPN \cite{kendall2016modelling} &  BMDN \cite{deng2022deep} & Ours\\
            \midrule
            %\multirow{7}{*}{\rotatebox[origin=c]{90}{\parbox[c]{1cm}{\centering 7-Scenes}}} & 
            Chess & $0.32/8.12$ & $\mathbf{0.08/3.25}$ & $0.37/7.24$ & $0.10/6.47$ & $0.17/6.90$\\
            Fire & $0.47/14.4$ & $0.27/\mathbf{11.7}$ & $0.43/13.7$ & $\mathbf{0.26}/14.8$ & $0.30/14.1$\\
            Heads & $0.29/\mathbf{12.0}$ & $0.18/13.3$ & $0.31/\mathbf{12.0}$ & $\mathbf{0.13}/13.4$ & $0.17/14.5$\\
            Office & $0.48/7.68$ & $\mathbf{0.17/5.15}$ & $0.48/8.04$ & $0.19/9.73$ & $0.24/9.30$\\
            Pumpkin & $0.47/8.42$ & $0.22/\mathbf{4.02}$ & $0.61/7.08$ & $\mathbf{0.20}/9.40$ & $0.30/8.33$\\
            Kitchen & $0.59/8.64$ & $0.23/\mathbf{4.93}$ & $0.58/7.54$ & $\mathbf{0.19}/10.9$ & $0.26/10.2$\\
            Stairs & $0.47/13.8$ & $\mathbf{0.30/12.1}$ & $0.48/13.1$ & $0.34/14.1$ & $0.47/15.5$\\
            \midrule
            %\multirow{5}{*}{\rotatebox[origin=c]{90}{\parbox[c]{1cm}{\centering Cambridge\\Landmarks}}} & 
            College & $1.92/5.40$ & $\mathbf{1.07/1.89}$ & $1.74/4.06$ & $1.51/2.14$ & $1.65/2.88$ \\
            Street & $3.67/6.50$ & $-$ & $\mathbf{2.96/6.00}$ & $16.3/25.2$ & $17.2/23.8$\\
            Hospital & $2.31/5.38$ & $\mathbf{1.94/3.91}$ & $2.57/5.14$ & $2.25/3.93$ & $2.06/4.33$\\
            Façade & $1.46/8.08$ & $1.49/\mathbf{4.22}$ & $1.25/7.54$ & $3.52/5.41$ & $\mathbf{1.02}/6.03$ \\
            Church & $2.65/8.48$ & $2.00/\mathbf{4.53}$ & $2.11/8.38$ & $2.16/5.99$ & $\mathbf{1.80}/5.90$\\
            \bottomrule
        \end{tabular}
        \begin{tablenotes}
             \item[$\dagger$] Results of MapNet on Cambridge Landmarks taken from Sattler et al. \cite{sattler2019understanding}.
        \end{tablenotes}
    \end{threeparttable}
\end{table}

For completeness, we report our results on the unambiguous 7-Scenes and Cambridge Landmarks datasets in Table \ref{tab:median_error}. We include results of PoseNet and MapNet as single pose regressor baselines, and Bayesian PoseNet and Bingham MDN as methods that, in principle, can predict multimodal distributions. While our method does not perform the best, it is not far from the top-performers. This experiment merely serves as a sanity check of our approach's performance in a minimal pipeline, without any particular mechanism aimed at improving accuracy in unambiguous scenes. As seen in Table \ref{tab:ambiguous}, the better performing methods on unambiguous scenes show poor performance on ambiguous scenes, which is the problem that our method targets to solve. An interesting direction for future work is to apply our proposed formulation, aimed at handling ambiguous scenes, in tandem with techniques for improved unambiguous pose regression.

% We can see that MapNet performs best in most of the scenes. Moreover, Bingham MDN shows lower errors than ours on some of the scenes. We also acknowledge that more recent works propose novel techniques that achieve state-of-the-art results on unambiguous datasets, outperforming our reported results \cite{chen2021direct, moreau2022lens, ng2021reassessing}.

\begin{figure}[t]
    \centering
    % SUBFLOAT
    \subfloat[Query image]{
        \includegraphics[width=0.13\textwidth]{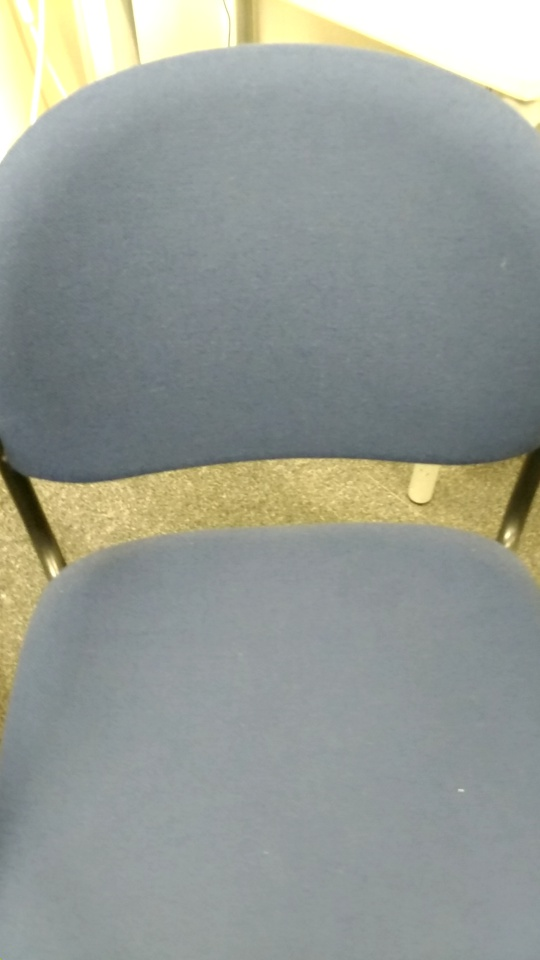}%
    }%
    \renewcommand{\arraystretch}{0}%
    \subfloat[Position posterior]{
        \begin{tabular}[b]{@{}c@{}}
            \resizebox{!}{0.13\textwidth}{%
                \begin{tikzpicture}%
                    \node[inner sep=0pt,outer sep=0pt,] (position) at (0,0) {\includegraphics[height=0.3\paperwidth]{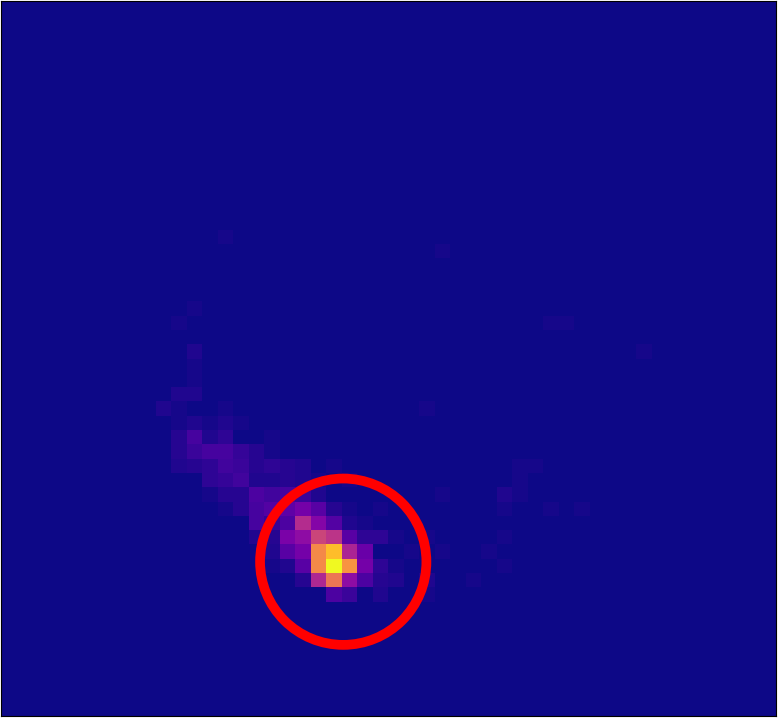}};%
                    % \node[inner sep=0pt,outer sep=0pt,anchor=west] (orientation) at ($(position.east) + (0.3, 0)$) {\includegraphics[height=0.22\paperwidth]{figures/squint_vs_accurate/accurate_orientation.png}};%
                \end{tikzpicture}%
            }\\\vspace{0.1cm}\\
            \resizebox{!}{0.13\textwidth}{%
                \begin{tikzpicture}
                    \node[inner sep=0pt,outer sep=0pt] (position) at (0,0) {\includegraphics[height=0.3\paperwidth]{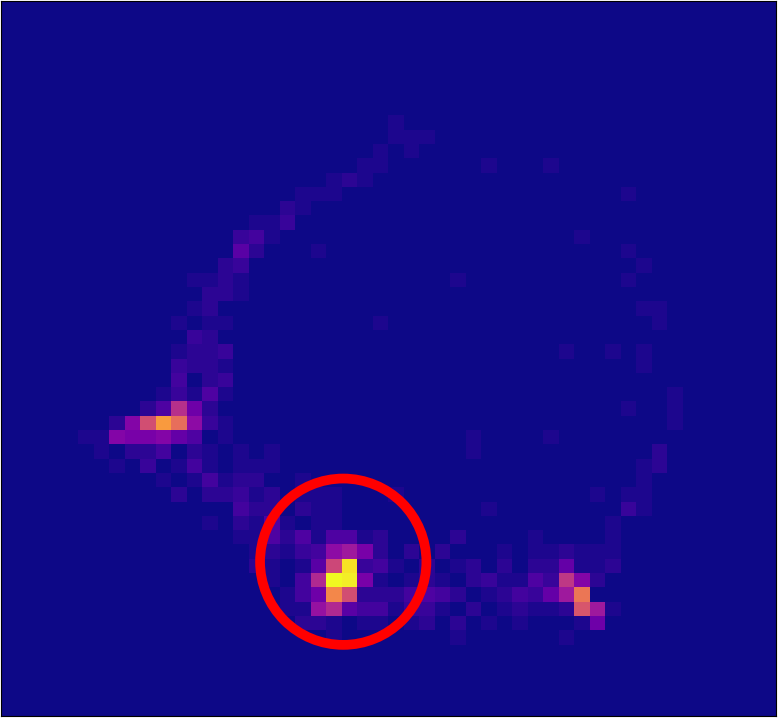}};%
                    % \node[inner sep=0pt,outer sep=0pt, anchor=west] (orientation) at ($(position.east) + (0.3, 0)$) {\includegraphics[height=0.22\paperwidth]{figures/squint_vs_accurate/squint_orientation.png}};%
                \end{tikzpicture}%
            }%
        \end{tabular}%
    }%
    \subfloat[Samples]{
        \begin{tabular}[b]{@{}c@{}}
            \includegraphics[trim={1cm 1cm 1.7cm 0.5cm},clip, height=0.13\textwidth]{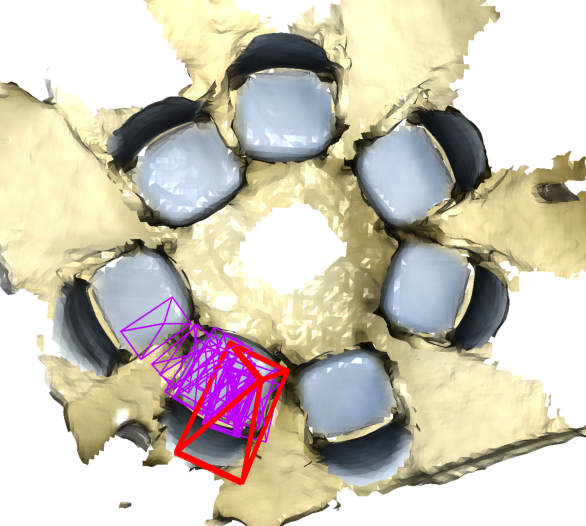}%
            \\\vspace{0.1cm}\\
            \includegraphics[trim={1cm 1cm 1.7cm 0.5cm},clip, height=0.13\textwidth]{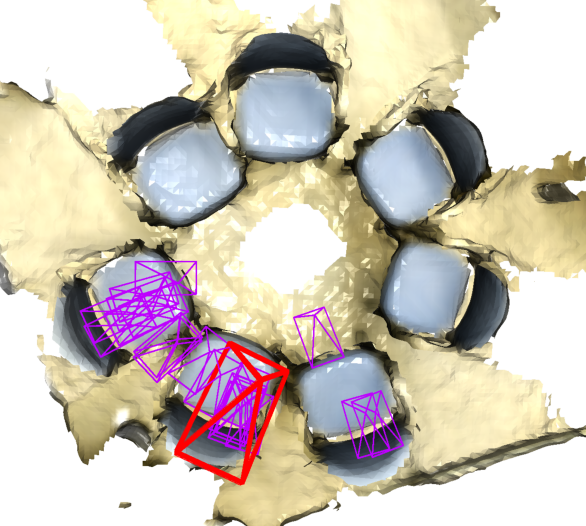}%
        \end{tabular}%
    }%
    \scriptsize
    \hspace{0.1cm}\rotatebox[origin=r]{-90}{No Weight Decay\hspace{0.85cm}Weight Decay\hspace{0.4cm}}
    \caption{Posteriors predicted by our method in full capacity (top) and in a constrained learning mode by $L_2$ weight decay of $\lambda = 0.1$ (bottom). The latter captures multiple modes whereas the former mode predicts a single mode at the correct pose.}\label{fig:squint}
\end{figure}

\subsection{A closer look}

Fig.~\ref{fig:squint} (top) shows the predicted distribution by our method for an example query image from the Ambiguous Relocalization dataset. Although the scene, made up of identical chairs, is arguably ambiguous to the human eye, we can see that the predicted posterior identifies and concentrates its density around the correct pose. We hypothesize that a sufficiently expressive \textit{Encoder} can distinguish a seemingly ambiguous image taken in real life by its smallest of details, such as the chair's background in this example. However, a less expressive \textit{Encoder} for the data is unable to learn every detail and can give in to the ambiguities. We test this hypothesis by adding a penalty term on the $L_2$ norm of the \textit{Encoder} weights during training. We can see in Fig.~\ref{fig:squint} (bottom) that this setting results in the predicted posterior assigning probabilities to poses viewing two additional chairs. We argue that when there exists a domain gap between the training data and the operation conditions, it is desirable for the model to trade off confidence in predictions for better generalization, which can be achieved via deliberate learning constraints. We leave the study of such learning constraints to future work.

\begin{figure}[t]
    \centering
    \scriptsize
    \setlength{\tabcolsep}{1.5pt}
    \setlength{\linewidth}{0.9\linewidth}
    \renewcommand{\arraystretch}{0}
    \begin{tabular}{lC{0.25\linewidth}C{0.25\linewidth}C{0.25\linewidth}C{0.25\linewidth}@{}}
        & Scene & $\alpha=0.01$ & $\alpha=0.20$ & $\alpha=1.00$\\\vspace{3pt}\\
        \rotatebox[origin=c]{90}{\parbox[c]{1.5cm}{\centering Round Table}}
        & \includegraphics[width=\linewidth]{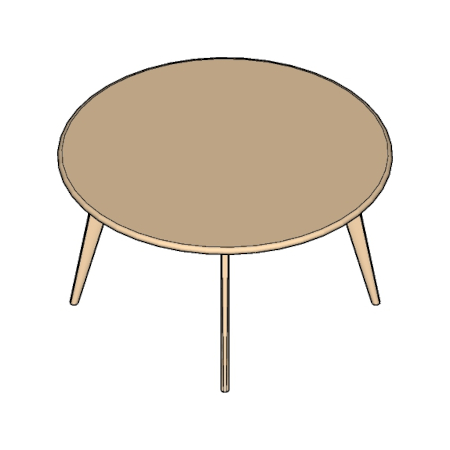} 
        & \includegraphics[width=\linewidth]{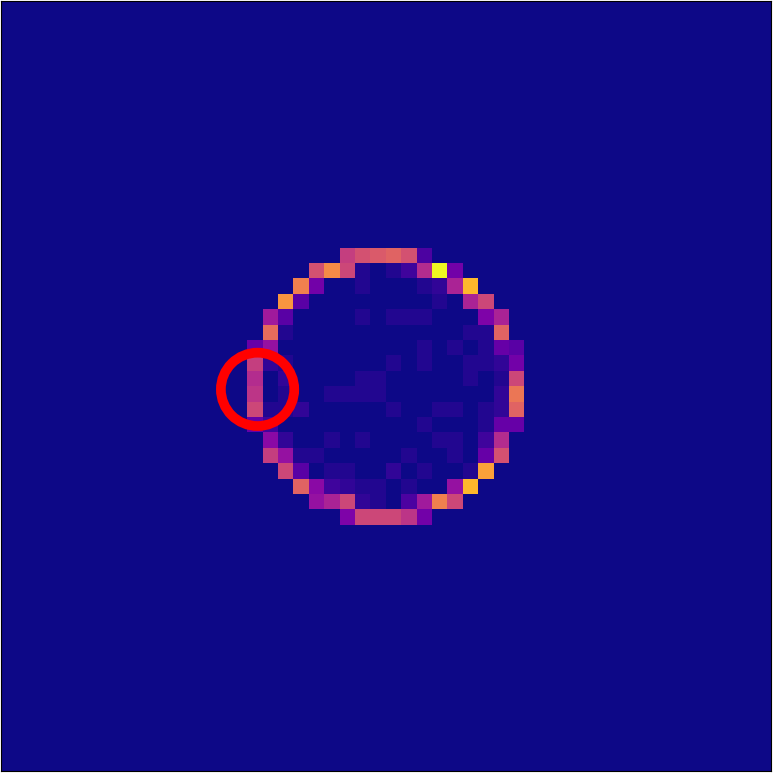}
        % & \includegraphics[width=\linewidth]{figures/synthetic_ablation/round_10.png}
        & \includegraphics[width=\linewidth]{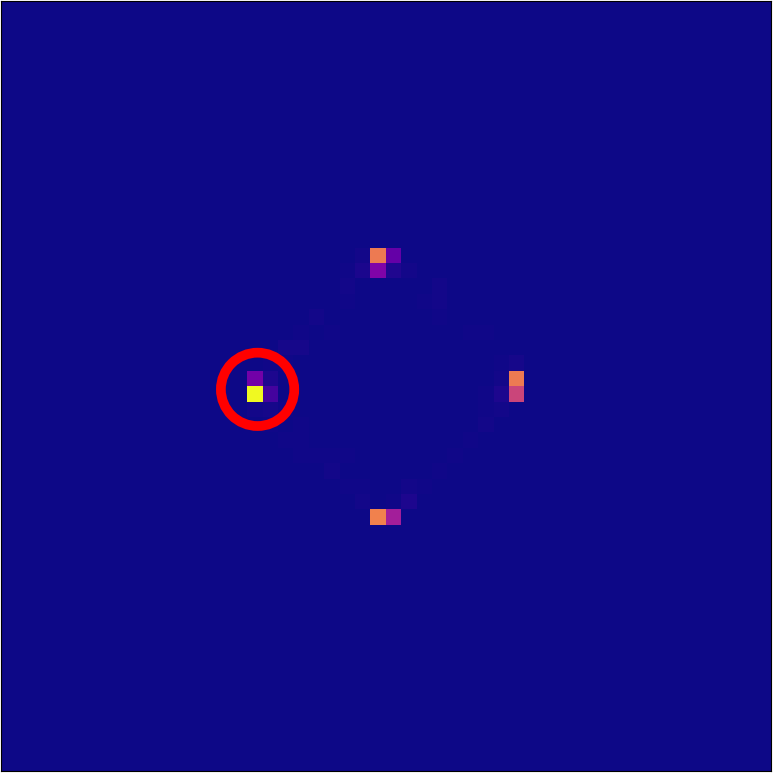}
        & \includegraphics[width=\linewidth]{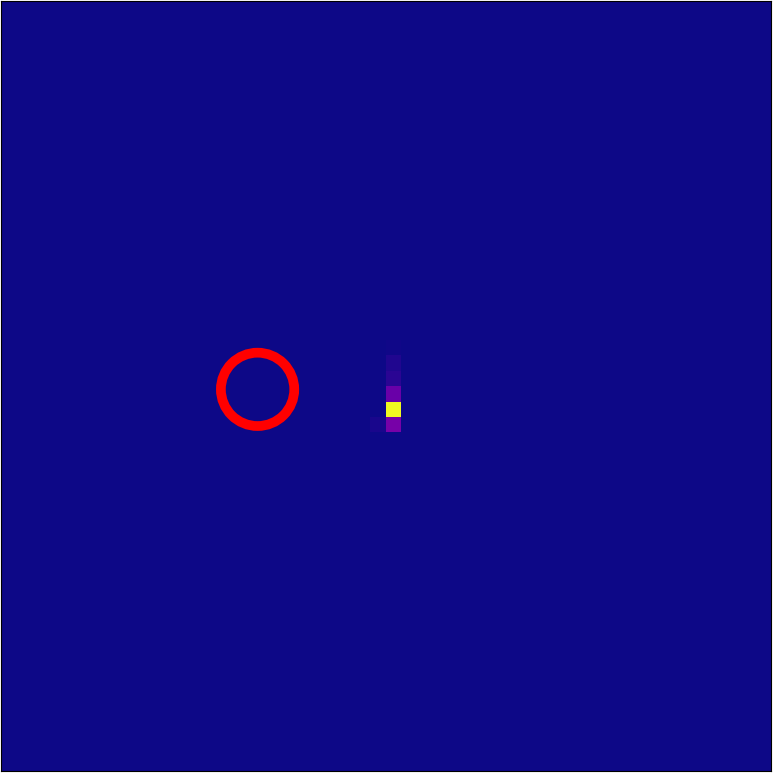}%
        \\\vspace{3pt}\\
        \rotatebox[origin=c]{90}{\parbox[c]{1.5cm}{\centering Dinner Table}}
        &\includegraphics[width=\linewidth]{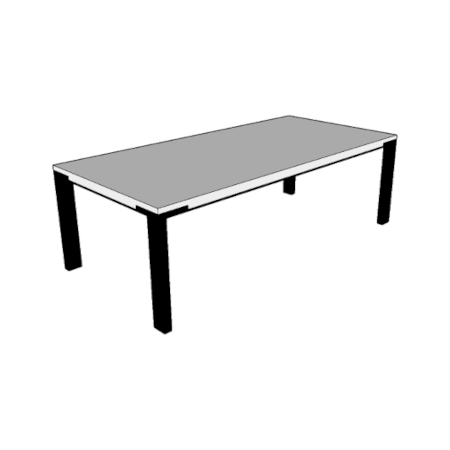}
        &\includegraphics[width=\linewidth]{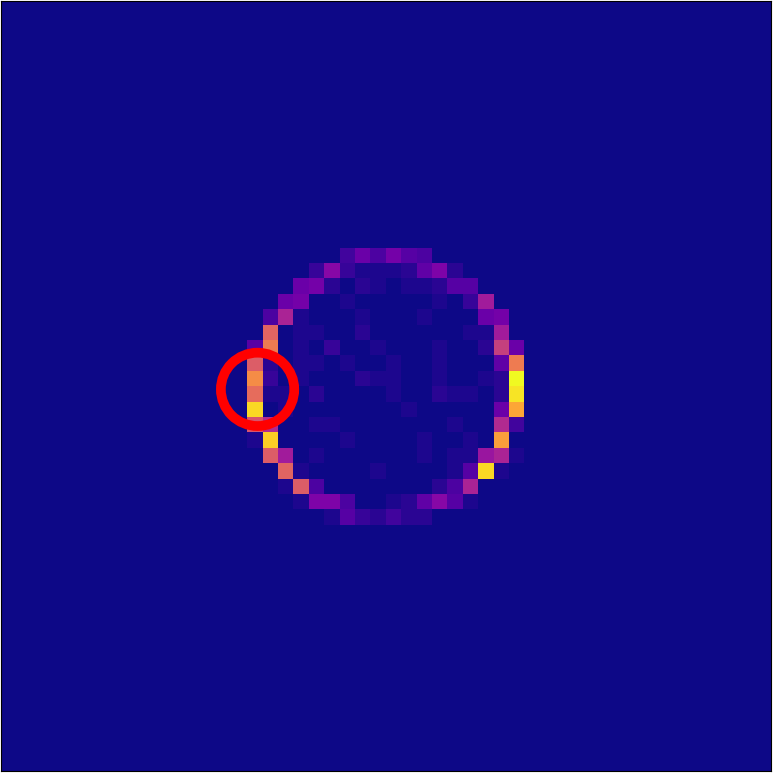}
        % &\includegraphics[width=\linewidth]{figures/synthetic_ablation/long_10.png}
        &\includegraphics[width=\linewidth]{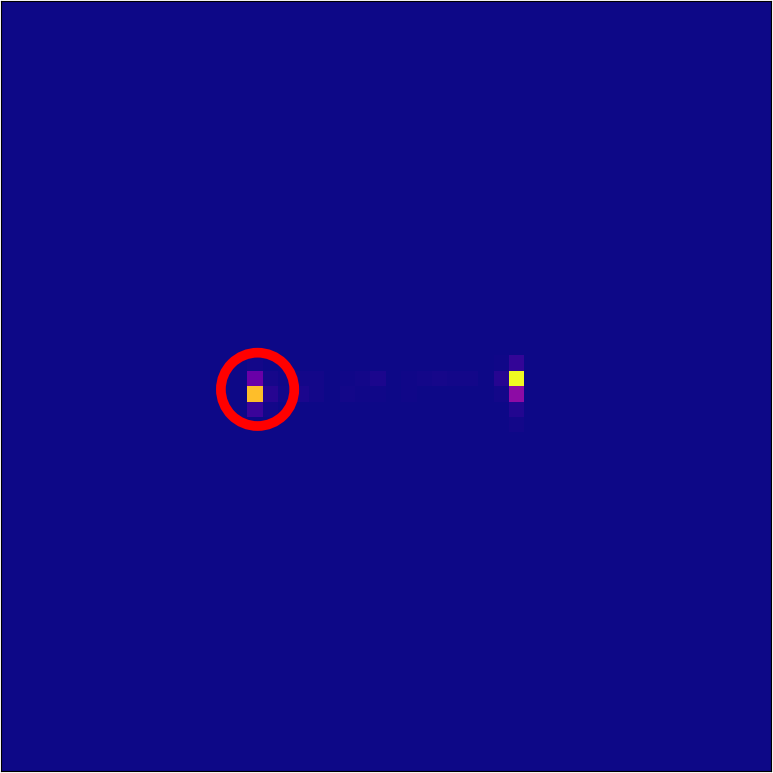}
        &\includegraphics[width=\linewidth]{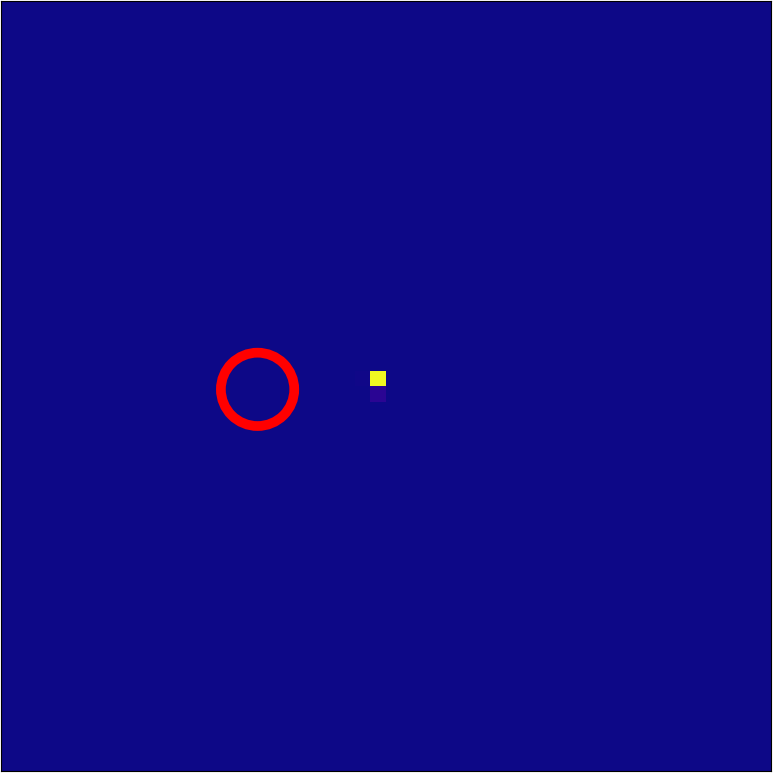}%
    \end{tabular}
    \caption{Predicted position posterior on $xy$-plane for various choices of $\alpha$ in the synthetic scenes. The round table and dinner table have four and two modes in their true distributions, respectively. The optimization successfully converges to predict the correct modes when $0 \ll \alpha < 1 / \#$ modes. For example $\alpha = 0.20$ produces good predictions for both scenes.}\label{fig:ablation}
\end{figure}

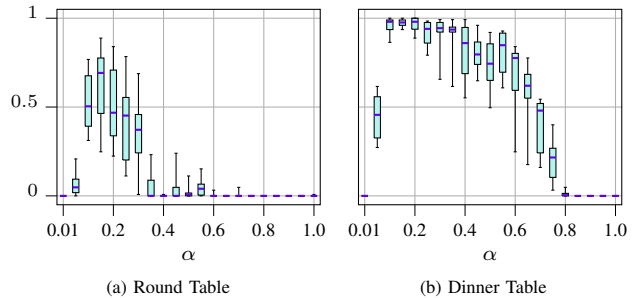
\begin{figure}[t]
    \centering
    \subfloat[Round Table]{%
        \centering%
        \subimport{figures/synthetic_ablation}{ablation_round_1000.tex}%
    }%
    \subfloat[Dinner Table]{%
        \centering%
        \subimport{figures/synthetic_ablation}{ablation_long_1000.tex}%
    }%
    \caption{Measured recall with threshold $0.1\text{m} / 10^{\circ}$ for various choices of $\alpha$ in Winners-Take-All optimization of the synthetic scenes. For each choice of $\alpha$, the statistics over 10 training runs is shown by a box extending from the lower to upper quartile recall values, a purple line at the median recall, and whiskers that extend to the minimum and maximum values.}\label{fig:ablationalpha}
\end{figure}

We study the effect of $\alpha$ in the Winners-Take-All optimization scheme in two synthetic scenes, where the camera circles around a round table with four legs, resulting in four modes in the pose distribution of an image, as well as a rectangular dinner table that results in bimodal distributions. We report the statistics over 10 training runs for the $0.1\text{m} / 10^{\circ}$ recall evaluated at the end of training with different $\alpha$ values in Fig.~\ref{fig:ablationalpha}. We can see that in these scenes the highest recall is achieved with $\alpha$ in a range of values greater than zero but less than $1/\#$ modes. Fig.~\ref{fig:ablation} shows the predicted camera position posterior for three choices of $\alpha$. We can see that a too large $\alpha$, as discussed in Section \ref{sec:problem_with_elbo}, results in a compromised posterior, and a too small $\alpha$ predicts close-to-uniform densities across the span of the training data. We hypothesize that $\alpha$ must be smaller than $1/\#$ modes for the Winners-Take-All optimization to converge and capture all modes in the distribution, and must be sufficiently larger than zero to overcome the noise as a result of mini-batch optimization. There is a trade-off between training speed and the quality of the learned distribution within this range of $\alpha$ values, as a smaller $\alpha$ results in optimization of fewer samples at every iteration, hence a slower training, but is less susceptible to the noise induced by Monte Carlo sampling. We leave the study of finding the optimal $\alpha$ to future work.

\subsection{Run-time evaluation}
We measure the time taken for a forward pass of one query image through our pipeline for 1000 Monte Carlo samples, on a desktop computer with an Intel Core i7-8700K CPU and an NVIDIA GeForce GTX 1080 Ti GPU. We repeat each measurement 100 times and we find that a forward pass on average takes $14.88 \pm 0.75\,$ms on CPU and $2.26 \pm 0.10\,$ms on GPU, that is, our pipeline can run in real time.

\section{Conclusion}
\label{sec:conclusion}

In this work, we addressed the task of visual localization in ambiguous scenes. We proposed a novel formulation of camera pose regression with variational inference, which allows learning and sampling from the distribution over all camera poses given an image. This is done by first encoding the images to predict a posterior distribution over the latent space of visual features present in the scene. Drawing samples from this distribution and passing them through a learned mapping produces a set of pose samples that represent the posterior distribution over camera poses. We show that our formulation outperforms existing methods on localization in ambiguous scenes, and propose directions for future work to further investigate our proposed method.

{\small
\bibliographystyle{IEEEtran}
\bibliography{egbib}
}

\end{document}

%% file: figures/pipeline.tex
\scriptsize

\begin{tikzpicture}[scale=1, thick]
    \coordinate (input_e) at (0,0);%
    \coordinate (encoder_c) at (1.2,0);%
    \coordinate (sampling_c) at (3.658,0);%
    \coordinate (posemap_c) at (5.9,0);%
    \coordinate (output_w) at (7.1,0);%
    
    % Input xi yi
    \node[draw=none, anchor=east] (x) at (input_e) {$\mathbf{x}$};
    \node[draw=none, anchor=east] (y) at ($(input_e) + (0.0,-0.5)$) {$y$};
    % \node[anchor=south west] (xiyionlyphantom) at (xiyi.south west) {$\phantom{\begin{aligned}
    %   \mathbf{x}\\ 
    %   y
    % \end{aligned}}$};
    
    % Prior
    \node[draw=none, anchor=west] (prior) at ($(x.west) + (0, 1.25)$) {$p(\mathbf{z}) = \mathcal{N}(\mathbf{z} ; \mathbf{0}, \mathbf{I})$};
    
    % Output yi
    \node[anchor=west] (output) at (output_w) {$\hat{y}$};
    \node[anchor=west] (yihatphantom) at (output_w) {$\phantom{\hat{y}}$};
    
    % Encoder
    % \node[align=center, draw, minimum width=2.5cm, minimum height=1.5cm] (encoder) at (encoder_c) {Encoder\\$q_\theta(\mathbf{z} \mid \mathbf{x}_i)$};
    \node[align=center, draw, minimum width=1.5cm, minimum height=1cm] (encoder) at (encoder_c) {Encoder\\$g_\theta(\cdot)$};
    
    % PoseMap
    \node[align=center, draw, minimum width=1.5cm, minimum height=1cm] (posemap) at (posemap_c) {PoseMap\\$f_\phi(\cdot)$};
    
    % Sampling
    % \node[align=center] (sampling) at (sampling_c) {$q(\mathbf{z} \mid \mathbf{x}) \backsim \mathbf{z}\in \mathbb{R}^d$};
    
    \newlength\simheight
    \settoheight\simheight{$\sim$}
    
    \node (sampling_circle) at (sampling_c) [inner sep=1pt, thick, anchor=center, draw=black, align=center, minimum height=0.5cm, circle] {\raisebox{0pt}[\simheight][0pt]{\raisebox{-1pt}{$\sim$}}};
    
    % \node[anchor=east] (qonlyphantom) at ($(sampling_circle.west) + (-0.2,0)$) {$q(\mathbf{z} \mid \mathbf{x})$};
    
    % \node[anchor=west] (zinrd) at ($(sampling_circle.east) + (0.2, 0)$) {$\mathbf{z}\in \mathbb{R}^d\vphantom{R^dq}$};

    % \pgfmathsetmacro\MathAxis{height("$\vcenter{}$")}
    % \node (a) at (0, 5) {%
    %   \sbox0{$=$}%
    %   \sbox0{\raisebox{-\MathAxis pt}{\usebox0}}% Equals sign at base line
    %   \dp0\sim\ht0 % The equals sign is symmetric
    %   \usebox0%
    % };
    % \node (b) at (1, 0) {$A=$};
    
    % Box
    \draw[materialteal, dashed] ($(posemap.north east) + (0.2,0.2)$) rectangle ($(encoder.south west) - (0.2,0.2)$);
    
    % Lines
    \draw[->] (x.east) -- (encoder.west);
    \draw[->] (encoder.east) -- node[above] (qzx) {$q(\mathbf{z} \mid \mathbf{x})$} (sampling_circle.west);
    \draw[->] (sampling_circle.east) -- node[above] {$\mathbf{z}\vphantom{R^dq}$} (posemap.west);
    % \draw[->] (qonlyphantom.east) -- (sampling_circle.west);
    % \draw[->] (sampling_circle.east) -- (zinrd.west);
    % \draw[->] (zinrd.east) -- (posemap.west);
    \draw[->] (posemap.east) -- (output.west);
    
    % % Geometric error line
    \draw[materialviolet,fill] (yihatphantom.south) circle (0.05);
    \draw[materialviolet,fill] (y.south) circle (0.05);
    \draw[materialviolet]  (yihatphantom.south) -- ++ (0, -0.9) node (temp) {} -- node[below] {Winners-Take-All Prediction Error}  (y |- temp)  -- (y.south);
    
    % KLD line
    \draw[materialviolet,fill] (prior.east) circle (0.05);
    \draw[materialviolet,fill] (qzx.north) circle (0.05);
    \draw[materialviolet] (prior.east) -- (qzx.north |- prior.west) -- (qzx);
    \node[anchor=north west,materialviolet,inner sep=0pt, outer sep=0pt] (kldlabel) at ($(qzx.north |- prior.west) + (0.1,0)$) {KL Divergence};

\end{tikzpicture}

%% file: figures/ceiling/007450.tex
\scriptsize

% This file was created with tikzplotlib v0.10.1.
\begin{tikzpicture}

\definecolor{darkgray176}{RGB}{176,176,176}
\definecolor{darkorange25512714}{RGB}{255,127,14}
\definecolor{lightgray204}{RGB}{204,204,204}
\definecolor{limegreen}{RGB}{50,205,50}
\definecolor{magenta}{RGB}{255,0,255}
\definecolor{steelblue31119180}{RGB}{31,119,180}

\begin{groupplot}[group style={group size=2 by 2, horizontal sep=2pt, vertical sep=2pt}]

\nextgroupplot[
legend cell align={left},
legend style={fill opacity=0.8, draw opacity=1, text opacity=1, draw=lightgray204},
tick align=outside,
tick pos=left,
x grid style={darkgray176},
xmin=0, xmax=16500,
xtick style={color=black},
xticklabels={},
ylabel={Density},
ymin=0, ymax=0.002,
scale only axis,
width=15cm,
height=1.136cm,
ticks=none,
xtick scale label code/.code={},
ytick scale label code/.code={},
xlabel near ticks,
ylabel near ticks,
hide axis,
]
\draw[draw=none,fill=steelblue31119180,fill opacity=0.9] (axis cs:0,0) rectangle (axis cs:100,0);
% \addlegendimage{ybar,ybar legend,draw=none,fill=steelblue31119180,fill opacity=0.9}
% \addlegendentry{Ours}

\draw[draw=none,fill=steelblue31119180,fill opacity=0.9] (axis cs:100,0) rectangle (axis cs:200,0);
\draw[draw=none,fill=steelblue31119180,fill opacity=0.9] (axis cs:200,0) rectangle (axis cs:300,0);
\draw[draw=none,fill=steelblue31119180,fill opacity=0.9] (axis cs:300,0) rectangle (axis cs:400,0);
\draw[draw=none,fill=steelblue31119180,fill opacity=0.9] (axis cs:400,0) rectangle (axis cs:500,0);
\draw[draw=none,fill=steelblue31119180,fill opacity=0.9] (axis cs:500,0) rectangle (axis cs:600,0);
\draw[draw=none,fill=steelblue31119180,fill opacity=0.9] (axis cs:600,0) rectangle (axis cs:700,2e-05);
\draw[draw=none,fill=steelblue31119180,fill opacity=0.9] (axis cs:700,0) rectangle (axis cs:800,0.0005);
\draw[draw=none,fill=steelblue31119180,fill opacity=0.9] (axis cs:800,0) rectangle (axis cs:900,0.00077);
\draw[draw=none,fill=steelblue31119180,fill opacity=0.9] (axis cs:900,0) rectangle (axis cs:1000,8e-05);
\draw[draw=none,fill=steelblue31119180,fill opacity=0.9] (axis cs:1000,0) rectangle (axis cs:1100,4e-05);
\draw[draw=none,fill=steelblue31119180,fill opacity=0.9] (axis cs:1100,0) rectangle (axis cs:1200,2e-05);
\draw[draw=none,fill=steelblue31119180,fill opacity=0.9] (axis cs:1200,0) rectangle (axis cs:1300,0);
\draw[draw=none,fill=steelblue31119180,fill opacity=0.9] (axis cs:1300,0) rectangle (axis cs:1400,1e-05);
\draw[draw=none,fill=steelblue31119180,fill opacity=0.9] (axis cs:1400,0) rectangle (axis cs:1500,0);
\draw[draw=none,fill=steelblue31119180,fill opacity=0.9] (axis cs:1500,0) rectangle (axis cs:1600,0);
\draw[draw=none,fill=steelblue31119180,fill opacity=0.9] (axis cs:1600,0) rectangle (axis cs:1700,0);
\draw[draw=none,fill=steelblue31119180,fill opacity=0.9] (axis cs:1700,0) rectangle (axis cs:1800,0);
\draw[draw=none,fill=steelblue31119180,fill opacity=0.9] (axis cs:1800,0) rectangle (axis cs:1900,0);
\draw[draw=none,fill=steelblue31119180,fill opacity=0.9] (axis cs:1900,0) rectangle (axis cs:2000,0);
\draw[draw=none,fill=steelblue31119180,fill opacity=0.9] (axis cs:2000,0) rectangle (axis cs:2100,0);
\draw[draw=none,fill=steelblue31119180,fill opacity=0.9] (axis cs:2100,0) rectangle (axis cs:2200,1e-05);
\draw[draw=none,fill=steelblue31119180,fill opacity=0.9] (axis cs:2200,0) rectangle (axis cs:2300,0);
\draw[draw=none,fill=steelblue31119180,fill opacity=0.9] (axis cs:2300,0) rectangle (axis cs:2400,0);
\draw[draw=none,fill=steelblue31119180,fill opacity=0.9] (axis cs:2400,0) rectangle (axis cs:2500,0);
\draw[draw=none,fill=steelblue31119180,fill opacity=0.9] (axis cs:2500,0) rectangle (axis cs:2600,1e-05);
\draw[draw=none,fill=steelblue31119180,fill opacity=0.9] (axis cs:2600,0) rectangle (axis cs:2700,0);
\draw[draw=none,fill=steelblue31119180,fill opacity=0.9] (axis cs:2700,0) rectangle (axis cs:2800,0);
\draw[draw=none,fill=steelblue31119180,fill opacity=0.9] (axis cs:2800,0) rectangle (axis cs:2900,0);
\draw[draw=none,fill=steelblue31119180,fill opacity=0.9] (axis cs:2900,0) rectangle (axis cs:3000,0);
\draw[draw=none,fill=steelblue31119180,fill opacity=0.9] (axis cs:3000,0) rectangle (axis cs:3100,0);
\draw[draw=none,fill=steelblue31119180,fill opacity=0.9] (axis cs:3100,0) rectangle (axis cs:3200,0);
\draw[draw=none,fill=steelblue31119180,fill opacity=0.9] (axis cs:3200,0) rectangle (axis cs:3300,0);
\draw[draw=none,fill=steelblue31119180,fill opacity=0.9] (axis cs:3300,0) rectangle (axis cs:3400,0);
\draw[draw=none,fill=steelblue31119180,fill opacity=0.9] (axis cs:3400,0) rectangle (axis cs:3500,0);
\draw[draw=none,fill=steelblue31119180,fill opacity=0.9] (axis cs:3500,0) rectangle (axis cs:3600,0);
\draw[draw=none,fill=steelblue31119180,fill opacity=0.9] (axis cs:3600,0) rectangle (axis cs:3700,0.00022);
\draw[draw=none,fill=steelblue31119180,fill opacity=0.9] (axis cs:3700,0) rectangle (axis cs:3800,0.00082);
\draw[draw=none,fill=steelblue31119180,fill opacity=0.9] (axis cs:3800,0) rectangle (axis cs:3900,0.0007);
\draw[draw=none,fill=steelblue31119180,fill opacity=0.9] (axis cs:3900,0) rectangle (axis cs:4000,0.0003);
\draw[draw=none,fill=steelblue31119180,fill opacity=0.9] (axis cs:4000,0) rectangle (axis cs:4100,2e-05);
\draw[draw=none,fill=steelblue31119180,fill opacity=0.9] (axis cs:4100,0) rectangle (axis cs:4200,0);
\draw[draw=none,fill=steelblue31119180,fill opacity=0.9] (axis cs:4200,0) rectangle (axis cs:4300,0);
\draw[draw=none,fill=steelblue31119180,fill opacity=0.9] (axis cs:4300,0) rectangle (axis cs:4400,0);
\draw[draw=none,fill=steelblue31119180,fill opacity=0.9] (axis cs:4400,0) rectangle (axis cs:4500,0);
\draw[draw=none,fill=steelblue31119180,fill opacity=0.9] (axis cs:4500,0) rectangle (axis cs:4600,0);
\draw[draw=none,fill=steelblue31119180,fill opacity=0.9] (axis cs:4600,0) rectangle (axis cs:4700,0);
\draw[draw=none,fill=steelblue31119180,fill opacity=0.9] (axis cs:4700,0) rectangle (axis cs:4800,1e-05);
\draw[draw=none,fill=steelblue31119180,fill opacity=0.9] (axis cs:4800,0) rectangle (axis cs:4900,0);
\draw[draw=none,fill=steelblue31119180,fill opacity=0.9] (axis cs:4900,0) rectangle (axis cs:5000,0);
\draw[draw=none,fill=steelblue31119180,fill opacity=0.9] (axis cs:5000,0) rectangle (axis cs:5100,0);
\draw[draw=none,fill=steelblue31119180,fill opacity=0.9] (axis cs:5100,0) rectangle (axis cs:5200,0);
\draw[draw=none,fill=steelblue31119180,fill opacity=0.9] (axis cs:5200,0) rectangle (axis cs:5300,0);
\draw[draw=none,fill=steelblue31119180,fill opacity=0.9] (axis cs:5300,0) rectangle (axis cs:5400,0);
\draw[draw=none,fill=steelblue31119180,fill opacity=0.9] (axis cs:5400,0) rectangle (axis cs:5500,1e-05);
\draw[draw=none,fill=steelblue31119180,fill opacity=0.9] (axis cs:5500,0) rectangle (axis cs:5600,0);
\draw[draw=none,fill=steelblue31119180,fill opacity=0.9] (axis cs:5600,0) rectangle (axis cs:5700,0);
\draw[draw=none,fill=steelblue31119180,fill opacity=0.9] (axis cs:5700,0) rectangle (axis cs:5800,1e-05);
\draw[draw=none,fill=steelblue31119180,fill opacity=0.9] (axis cs:5800,0) rectangle (axis cs:5900,0);
\draw[draw=none,fill=steelblue31119180,fill opacity=0.9] (axis cs:5900,0) rectangle (axis cs:6000,1e-05);
\draw[draw=none,fill=steelblue31119180,fill opacity=0.9] (axis cs:6000,0) rectangle (axis cs:6100,0);
\draw[draw=none,fill=steelblue31119180,fill opacity=0.9] (axis cs:6100,0) rectangle (axis cs:6200,1e-05);
\draw[draw=none,fill=steelblue31119180,fill opacity=0.9] (axis cs:6200,0) rectangle (axis cs:6300,1e-05);
\draw[draw=none,fill=steelblue31119180,fill opacity=0.9] (axis cs:6300,0) rectangle (axis cs:6400,0);
\draw[draw=none,fill=steelblue31119180,fill opacity=0.9] (axis cs:6400,0) rectangle (axis cs:6500,1e-05);
\draw[draw=none,fill=steelblue31119180,fill opacity=0.9] (axis cs:6500,0) rectangle (axis cs:6600,0);
\draw[draw=none,fill=steelblue31119180,fill opacity=0.9] (axis cs:6600,0) rectangle (axis cs:6700,4e-05);
\draw[draw=none,fill=steelblue31119180,fill opacity=0.9] (axis cs:6700,0) rectangle (axis cs:6800,0.00125);
\draw[draw=none,fill=steelblue31119180,fill opacity=0.9] (axis cs:6800,0) rectangle (axis cs:6900,0.00059);
\draw[draw=none,fill=steelblue31119180,fill opacity=0.9] (axis cs:6900,0) rectangle (axis cs:7000,2e-05);
\draw[draw=none,fill=steelblue31119180,fill opacity=0.9] (axis cs:7000,0) rectangle (axis cs:7100,3e-05);
\draw[draw=none,fill=steelblue31119180,fill opacity=0.9] (axis cs:7100,0) rectangle (axis cs:7200,0);
\draw[draw=none,fill=steelblue31119180,fill opacity=0.9] (axis cs:7200,0) rectangle (axis cs:7300,0);
\draw[draw=none,fill=steelblue31119180,fill opacity=0.9] (axis cs:7300,0) rectangle (axis cs:7400,0);
\draw[draw=none,fill=steelblue31119180,fill opacity=0.9] (axis cs:7400,0) rectangle (axis cs:7500,0);
\draw[draw=none,fill=steelblue31119180,fill opacity=0.9] (axis cs:7500,0) rectangle (axis cs:7600,0);
\draw[draw=none,fill=steelblue31119180,fill opacity=0.9] (axis cs:7600,0) rectangle (axis cs:7700,3e-05);
\draw[draw=none,fill=steelblue31119180,fill opacity=0.9] (axis cs:7700,0) rectangle (axis cs:7800,1e-05);
\draw[draw=none,fill=steelblue31119180,fill opacity=0.9] (axis cs:7800,0) rectangle (axis cs:7900,0);
\draw[draw=none,fill=steelblue31119180,fill opacity=0.9] (axis cs:7900,0) rectangle (axis cs:8000,0);
\draw[draw=none,fill=steelblue31119180,fill opacity=0.9] (axis cs:8000,0) rectangle (axis cs:8100,0);
\draw[draw=none,fill=steelblue31119180,fill opacity=0.9] (axis cs:8100,0) rectangle (axis cs:8200,2e-05);
\draw[draw=none,fill=steelblue31119180,fill opacity=0.9] (axis cs:8200,0) rectangle (axis cs:8300,1e-05);
\draw[draw=none,fill=steelblue31119180,fill opacity=0.9] (axis cs:8300,0) rectangle (axis cs:8400,0);
\draw[draw=none,fill=steelblue31119180,fill opacity=0.9] (axis cs:8400,0) rectangle (axis cs:8500,0);
\draw[draw=none,fill=steelblue31119180,fill opacity=0.9] (axis cs:8500,0) rectangle (axis cs:8600,0);
\draw[draw=none,fill=steelblue31119180,fill opacity=0.9] (axis cs:8600,0) rectangle (axis cs:8700,2e-05);
\draw[draw=none,fill=steelblue31119180,fill opacity=0.9] (axis cs:8700,0) rectangle (axis cs:8800,0);
\draw[draw=none,fill=steelblue31119180,fill opacity=0.9] (axis cs:8800,0) rectangle (axis cs:8900,0);
\draw[draw=none,fill=steelblue31119180,fill opacity=0.9] (axis cs:8900,0) rectangle (axis cs:9000,0);
\draw[draw=none,fill=steelblue31119180,fill opacity=0.9] (axis cs:9000,0) rectangle (axis cs:9100,1e-05);
\draw[draw=none,fill=steelblue31119180,fill opacity=0.9] (axis cs:9100,0) rectangle (axis cs:9200,0);
\draw[draw=none,fill=steelblue31119180,fill opacity=0.9] (axis cs:9200,0) rectangle (axis cs:9300,3e-05);
\draw[draw=none,fill=steelblue31119180,fill opacity=0.9] (axis cs:9300,0) rectangle (axis cs:9400,3e-05);
\draw[draw=none,fill=steelblue31119180,fill opacity=0.9] (axis cs:9400,0) rectangle (axis cs:9500,0);
\draw[draw=none,fill=steelblue31119180,fill opacity=0.9] (axis cs:9500,0) rectangle (axis cs:9600,2e-05);
\draw[draw=none,fill=steelblue31119180,fill opacity=0.9] (axis cs:9600,0) rectangle (axis cs:9700,0.00046);
\draw[draw=none,fill=steelblue31119180,fill opacity=0.9] (axis cs:9700,0) rectangle (axis cs:9800,0.00097);
\draw[draw=none,fill=steelblue31119180,fill opacity=0.9] (axis cs:9800,0) rectangle (axis cs:9900,6e-05);
\draw[draw=none,fill=steelblue31119180,fill opacity=0.9] (axis cs:9900,0) rectangle (axis cs:10000,9e-05);
\draw[draw=none,fill=steelblue31119180,fill opacity=0.9] (axis cs:10000,0) rectangle (axis cs:10100,0.00019);
\draw[draw=none,fill=steelblue31119180,fill opacity=0.9] (axis cs:10100,0) rectangle (axis cs:10200,0.00012);
\draw[draw=none,fill=steelblue31119180,fill opacity=0.9] (axis cs:10200,0) rectangle (axis cs:10300,1e-05);
\draw[draw=none,fill=steelblue31119180,fill opacity=0.9] (axis cs:10300,0) rectangle (axis cs:10400,0);
\draw[draw=none,fill=steelblue31119180,fill opacity=0.9] (axis cs:10400,0) rectangle (axis cs:10500,0);
\draw[draw=none,fill=steelblue31119180,fill opacity=0.9] (axis cs:10500,0) rectangle (axis cs:10600,0);
\draw[draw=none,fill=steelblue31119180,fill opacity=0.9] (axis cs:10600,0) rectangle (axis cs:10700,0);
\draw[draw=none,fill=steelblue31119180,fill opacity=0.9] (axis cs:10700,0) rectangle (axis cs:10800,0);
\draw[draw=none,fill=steelblue31119180,fill opacity=0.9] (axis cs:10800,0) rectangle (axis cs:10900,0);
\draw[draw=none,fill=steelblue31119180,fill opacity=0.9] (axis cs:10900,0) rectangle (axis cs:11000,0);
\draw[draw=none,fill=steelblue31119180,fill opacity=0.9] (axis cs:11000,0) rectangle (axis cs:11100,0);
\draw[draw=none,fill=steelblue31119180,fill opacity=0.9] (axis cs:11100,0) rectangle (axis cs:11200,0);
\draw[draw=none,fill=steelblue31119180,fill opacity=0.9] (axis cs:11200,0) rectangle (axis cs:11300,0);
\draw[draw=none,fill=steelblue31119180,fill opacity=0.9] (axis cs:11300,0) rectangle (axis cs:11400,0);
\draw[draw=none,fill=steelblue31119180,fill opacity=0.9] (axis cs:11400,0) rectangle (axis cs:11500,0);
\draw[draw=none,fill=steelblue31119180,fill opacity=0.9] (axis cs:11500,0) rectangle (axis cs:11600,0);
\draw[draw=none,fill=steelblue31119180,fill opacity=0.9] (axis cs:11600,0) rectangle (axis cs:11700,0);
\draw[draw=none,fill=steelblue31119180,fill opacity=0.9] (axis cs:11700,0) rectangle (axis cs:11800,0);
\draw[draw=none,fill=steelblue31119180,fill opacity=0.9] (axis cs:11800,0) rectangle (axis cs:11900,0);
\draw[draw=none,fill=steelblue31119180,fill opacity=0.9] (axis cs:11900,0) rectangle (axis cs:12000,0);
\draw[draw=none,fill=steelblue31119180,fill opacity=0.9] (axis cs:12000,0) rectangle (axis cs:12100,0);
\draw[draw=none,fill=steelblue31119180,fill opacity=0.9] (axis cs:12100,0) rectangle (axis cs:12200,0);
\draw[draw=none,fill=steelblue31119180,fill opacity=0.9] (axis cs:12200,0) rectangle (axis cs:12300,0);
\draw[draw=none,fill=steelblue31119180,fill opacity=0.9] (axis cs:12300,0) rectangle (axis cs:12400,0);
\draw[draw=none,fill=steelblue31119180,fill opacity=0.9] (axis cs:12400,0) rectangle (axis cs:12500,0);
\draw[draw=none,fill=steelblue31119180,fill opacity=0.9] (axis cs:12500,0) rectangle (axis cs:12600,0);
\draw[draw=none,fill=steelblue31119180,fill opacity=0.9] (axis cs:12600,0) rectangle (axis cs:12700,1e-05);
\draw[draw=none,fill=steelblue31119180,fill opacity=0.9] (axis cs:12700,0) rectangle (axis cs:12800,6e-05);
\draw[draw=none,fill=steelblue31119180,fill opacity=0.9] (axis cs:12800,0) rectangle (axis cs:12900,0.00083);
\draw[draw=none,fill=steelblue31119180,fill opacity=0.9] (axis cs:12900,0) rectangle (axis cs:13000,0.00035);
\draw[draw=none,fill=steelblue31119180,fill opacity=0.9] (axis cs:13000,0) rectangle (axis cs:13100,0.00024);
\draw[draw=none,fill=steelblue31119180,fill opacity=0.9] (axis cs:13100,0) rectangle (axis cs:13200,3e-05);
\draw[draw=none,fill=steelblue31119180,fill opacity=0.9] (axis cs:13200,0) rectangle (axis cs:13300,0);
\draw[draw=none,fill=steelblue31119180,fill opacity=0.9] (axis cs:13300,0) rectangle (axis cs:13400,0);
\draw[draw=none,fill=steelblue31119180,fill opacity=0.9] (axis cs:13400,0) rectangle (axis cs:13500,0);
\draw[draw=none,fill=steelblue31119180,fill opacity=0.9] (axis cs:13500,0) rectangle (axis cs:13600,0);
\draw[draw=none,fill=steelblue31119180,fill opacity=0.9] (axis cs:13600,0) rectangle (axis cs:13700,1e-05);
\draw[draw=none,fill=steelblue31119180,fill opacity=0.9] (axis cs:13700,0) rectangle (axis cs:13800,0);
\draw[draw=none,fill=steelblue31119180,fill opacity=0.9] (axis cs:13800,0) rectangle (axis cs:13900,0);
\draw[draw=none,fill=steelblue31119180,fill opacity=0.9] (axis cs:13900,0) rectangle (axis cs:14000,0);
\draw[draw=none,fill=steelblue31119180,fill opacity=0.9] (axis cs:14000,0) rectangle (axis cs:14100,0);
\draw[draw=none,fill=steelblue31119180,fill opacity=0.9] (axis cs:14100,0) rectangle (axis cs:14200,0);
\draw[draw=none,fill=steelblue31119180,fill opacity=0.9] (axis cs:14200,0) rectangle (axis cs:14300,0);
\draw[draw=none,fill=steelblue31119180,fill opacity=0.9] (axis cs:14300,0) rectangle (axis cs:14400,0);
\draw[draw=none,fill=steelblue31119180,fill opacity=0.9] (axis cs:14400,0) rectangle (axis cs:14500,0);
\draw[draw=none,fill=steelblue31119180,fill opacity=0.9] (axis cs:14500,0) rectangle (axis cs:14600,0);
\draw[draw=none,fill=steelblue31119180,fill opacity=0.9] (axis cs:14600,0) rectangle (axis cs:14700,0);
\draw[draw=none,fill=steelblue31119180,fill opacity=0.9] (axis cs:14700,0) rectangle (axis cs:14800,0);
\draw[draw=none,fill=steelblue31119180,fill opacity=0.9] (axis cs:14800,0) rectangle (axis cs:14900,0);
\draw[draw=none,fill=steelblue31119180,fill opacity=0.9] (axis cs:14900,0) rectangle (axis cs:15000,1e-05);
\draw[draw=none,fill=steelblue31119180,fill opacity=0.9] (axis cs:15000,0) rectangle (axis cs:15100,1e-05);
\draw[draw=none,fill=steelblue31119180,fill opacity=0.9] (axis cs:15100,0) rectangle (axis cs:15200,0);
\draw[draw=none,fill=steelblue31119180,fill opacity=0.9] (axis cs:15200,0) rectangle (axis cs:15300,1e-05);
\draw[draw=none,fill=steelblue31119180,fill opacity=0.9] (axis cs:15300,0) rectangle (axis cs:15400,0);
\draw[draw=none,fill=steelblue31119180,fill opacity=0.9] (axis cs:15400,0) rectangle (axis cs:15500,1e-05);
\draw[draw=none,fill=steelblue31119180,fill opacity=0.9] (axis cs:15500,0) rectangle (axis cs:15600,8e-05);
\draw[draw=none,fill=steelblue31119180,fill opacity=0.9] (axis cs:15600,0) rectangle (axis cs:15700,0.00023);
\draw[draw=none,fill=steelblue31119180,fill opacity=0.9] (axis cs:15700,0) rectangle (axis cs:15800,0.0002);
\draw[draw=none,fill=steelblue31119180,fill opacity=0.9] (axis cs:15800,0) rectangle (axis cs:15900,0.00025);
\draw[draw=none,fill=steelblue31119180,fill opacity=0.9] (axis cs:15900,0) rectangle (axis cs:16000,7e-05);
\draw[draw=none,fill=steelblue31119180,fill opacity=0.9] (axis cs:16000,0) rectangle (axis cs:16100,0);
\draw[draw=none,fill=steelblue31119180,fill opacity=0.9] (axis cs:16100,0) rectangle (axis cs:16200,0);
\draw[draw=none,fill=steelblue31119180,fill opacity=0.9] (axis cs:16200,0) rectangle (axis cs:16300,0);
\draw[draw=none,fill=steelblue31119180,fill opacity=0.9] (axis cs:16300,0) rectangle (axis cs:16400,0);
\draw[draw=none,fill=steelblue31119180,fill opacity=0.9] (axis cs:16400,0) rectangle (axis cs:16500,0);
\path [draw=magenta, thick]
(axis cs:4045.08673095703,0)
--(axis cs:4045.08673095703,0.002);

\draw[draw=none,fill=darkorange25512714,fill opacity=0.7] (axis cs:0,0) rectangle (axis cs:100,0);
% \addlegendimage{ybar,ybar legend,draw=none,fill=darkorange25512714,fill opacity=0.7}
% \addlegendentry{Bui et al.}

\draw[draw=none,fill=darkorange25512714,fill opacity=0.7] (axis cs:100,0) rectangle (axis cs:200,1.00200400801603e-05);
\draw[draw=none,fill=darkorange25512714,fill opacity=0.7] (axis cs:200,0) rectangle (axis cs:300,0);
\draw[draw=none,fill=darkorange25512714,fill opacity=0.7] (axis cs:300,0) rectangle (axis cs:400,0);
\draw[draw=none,fill=darkorange25512714,fill opacity=0.7] (axis cs:400,0) rectangle (axis cs:500,0);
\draw[draw=none,fill=darkorange25512714,fill opacity=0.7] (axis cs:500,0) rectangle (axis cs:600,0);
\draw[draw=none,fill=darkorange25512714,fill opacity=0.7] (axis cs:600,0) rectangle (axis cs:700,0);
\draw[draw=none,fill=darkorange25512714,fill opacity=0.7] (axis cs:700,0) rectangle (axis cs:800,0);
\draw[draw=none,fill=darkorange25512714,fill opacity=0.7] (axis cs:800,0) rectangle (axis cs:900,0);
\draw[draw=none,fill=darkorange25512714,fill opacity=0.7] (axis cs:900,0) rectangle (axis cs:1000,0);
\draw[draw=none,fill=darkorange25512714,fill opacity=0.7] (axis cs:1000,0) rectangle (axis cs:1100,0);
\draw[draw=none,fill=darkorange25512714,fill opacity=0.7] (axis cs:1100,0) rectangle (axis cs:1200,0);
\draw[draw=none,fill=darkorange25512714,fill opacity=0.7] (axis cs:1200,0) rectangle (axis cs:1300,0);
\draw[draw=none,fill=darkorange25512714,fill opacity=0.7] (axis cs:1300,0) rectangle (axis cs:1400,0);
\draw[draw=none,fill=darkorange25512714,fill opacity=0.7] (axis cs:1400,0) rectangle (axis cs:1500,2.00400801603206e-05);
\draw[draw=none,fill=darkorange25512714,fill opacity=0.7] (axis cs:1500,0) rectangle (axis cs:1600,9.01803607214429e-05);
\draw[draw=none,fill=darkorange25512714,fill opacity=0.7] (axis cs:1600,0) rectangle (axis cs:1700,0.000230460921843687);
\draw[draw=none,fill=darkorange25512714,fill opacity=0.7] (axis cs:1700,0) rectangle (axis cs:1800,0.000380761523046092);
\draw[draw=none,fill=darkorange25512714,fill opacity=0.7] (axis cs:1800,0) rectangle (axis cs:1900,0.000190380761523046);
\draw[draw=none,fill=darkorange25512714,fill opacity=0.7] (axis cs:1900,0) rectangle (axis cs:2000,0.000240480961923848);
\draw[draw=none,fill=darkorange25512714,fill opacity=0.7] (axis cs:2000,0) rectangle (axis cs:2100,0.000290581162324649);
\draw[draw=none,fill=darkorange25512714,fill opacity=0.7] (axis cs:2100,0) rectangle (axis cs:2200,0.000200400801603206);
\draw[draw=none,fill=darkorange25512714,fill opacity=0.7] (axis cs:2200,0) rectangle (axis cs:2300,7.01402805611223e-05);
\draw[draw=none,fill=darkorange25512714,fill opacity=0.7] (axis cs:2300,0) rectangle (axis cs:2400,2.00400801603206e-05);
\draw[draw=none,fill=darkorange25512714,fill opacity=0.7] (axis cs:2400,0) rectangle (axis cs:2500,1.00200400801603e-05);
\draw[draw=none,fill=darkorange25512714,fill opacity=0.7] (axis cs:2500,0) rectangle (axis cs:2600,1.00200400801603e-05);
\draw[draw=none,fill=darkorange25512714,fill opacity=0.7] (axis cs:2600,0) rectangle (axis cs:2700,0);
\draw[draw=none,fill=darkorange25512714,fill opacity=0.7] (axis cs:2700,0) rectangle (axis cs:2800,3.0060120240481e-05);
\draw[draw=none,fill=darkorange25512714,fill opacity=0.7] (axis cs:2800,0) rectangle (axis cs:2900,0);
\draw[draw=none,fill=darkorange25512714,fill opacity=0.7] (axis cs:2900,0) rectangle (axis cs:3000,2.00400801603206e-05);
\draw[draw=none,fill=darkorange25512714,fill opacity=0.7] (axis cs:3000,0) rectangle (axis cs:3100,0);
\draw[draw=none,fill=darkorange25512714,fill opacity=0.7] (axis cs:3100,0) rectangle (axis cs:3200,0);
\draw[draw=none,fill=darkorange25512714,fill opacity=0.7] (axis cs:3200,0) rectangle (axis cs:3300,0);
\draw[draw=none,fill=darkorange25512714,fill opacity=0.7] (axis cs:3300,0) rectangle (axis cs:3400,3.0060120240481e-05);
\draw[draw=none,fill=darkorange25512714,fill opacity=0.7] (axis cs:3400,0) rectangle (axis cs:3500,1.00200400801603e-05);
\draw[draw=none,fill=darkorange25512714,fill opacity=0.7] (axis cs:3500,0) rectangle (axis cs:3600,1.00200400801603e-05);
\draw[draw=none,fill=darkorange25512714,fill opacity=0.7] (axis cs:3600,0) rectangle (axis cs:3700,1.00200400801603e-05);
\draw[draw=none,fill=darkorange25512714,fill opacity=0.7] (axis cs:3700,0) rectangle (axis cs:3800,0);
\draw[draw=none,fill=darkorange25512714,fill opacity=0.7] (axis cs:3800,0) rectangle (axis cs:3900,2.00400801603206e-05);
\draw[draw=none,fill=darkorange25512714,fill opacity=0.7] (axis cs:3900,0) rectangle (axis cs:4000,1.00200400801603e-05);
\draw[draw=none,fill=darkorange25512714,fill opacity=0.7] (axis cs:4000,0) rectangle (axis cs:4100,0);
\draw[draw=none,fill=darkorange25512714,fill opacity=0.7] (axis cs:4100,0) rectangle (axis cs:4200,1.00200400801603e-05);
\draw[draw=none,fill=darkorange25512714,fill opacity=0.7] (axis cs:4200,0) rectangle (axis cs:4300,1.00200400801603e-05);
\draw[draw=none,fill=darkorange25512714,fill opacity=0.7] (axis cs:4300,0) rectangle (axis cs:4400,2.00400801603206e-05);
\draw[draw=none,fill=darkorange25512714,fill opacity=0.7] (axis cs:4400,0) rectangle (axis cs:4500,2.00400801603206e-05);
\draw[draw=none,fill=darkorange25512714,fill opacity=0.7] (axis cs:4500,0) rectangle (axis cs:4600,4.00801603206413e-05);
\draw[draw=none,fill=darkorange25512714,fill opacity=0.7] (axis cs:4600,0) rectangle (axis cs:4700,2.00400801603206e-05);
\draw[draw=none,fill=darkorange25512714,fill opacity=0.7] (axis cs:4700,0) rectangle (axis cs:4800,2.00400801603206e-05);
\draw[draw=none,fill=darkorange25512714,fill opacity=0.7] (axis cs:4800,0) rectangle (axis cs:4900,3.0060120240481e-05);
\draw[draw=none,fill=darkorange25512714,fill opacity=0.7] (axis cs:4900,0) rectangle (axis cs:5000,2.00400801603206e-05);
\draw[draw=none,fill=darkorange25512714,fill opacity=0.7] (axis cs:5000,0) rectangle (axis cs:5100,0.000230460921843687);
\draw[draw=none,fill=darkorange25512714,fill opacity=0.7] (axis cs:5100,0) rectangle (axis cs:5200,0.000410821643286573);
\draw[draw=none,fill=darkorange25512714,fill opacity=0.7] (axis cs:5200,0) rectangle (axis cs:5300,0.000270541082164329);
\draw[draw=none,fill=darkorange25512714,fill opacity=0.7] (axis cs:5300,0) rectangle (axis cs:5400,8.01603206412826e-05);
\draw[draw=none,fill=darkorange25512714,fill opacity=0.7] (axis cs:5400,0) rectangle (axis cs:5500,0.000561122244488978);
\draw[draw=none,fill=darkorange25512714,fill opacity=0.7] (axis cs:5500,0) rectangle (axis cs:5600,0.000370741482965932);
\draw[draw=none,fill=darkorange25512714,fill opacity=0.7] (axis cs:5600,0) rectangle (axis cs:5700,1.00200400801603e-05);
\draw[draw=none,fill=darkorange25512714,fill opacity=0.7] (axis cs:5700,0) rectangle (axis cs:5800,3.0060120240481e-05);
\draw[draw=none,fill=darkorange25512714,fill opacity=0.7] (axis cs:5800,0) rectangle (axis cs:5900,9.01803607214429e-05);
\draw[draw=none,fill=darkorange25512714,fill opacity=0.7] (axis cs:5900,0) rectangle (axis cs:6000,0.00146292585170341);
\draw[draw=none,fill=darkorange25512714,fill opacity=0.7] (axis cs:6000,0) rectangle (axis cs:6100,0.0011623246492986);
\draw[draw=none,fill=darkorange25512714,fill opacity=0.7] (axis cs:6100,0) rectangle (axis cs:6200,0.000160320641282565);
\draw[draw=none,fill=darkorange25512714,fill opacity=0.7] (axis cs:6200,0) rectangle (axis cs:6300,2.00400801603206e-05);
\draw[draw=none,fill=darkorange25512714,fill opacity=0.7] (axis cs:6300,0) rectangle (axis cs:6400,3.0060120240481e-05);
\draw[draw=none,fill=darkorange25512714,fill opacity=0.7] (axis cs:6400,0) rectangle (axis cs:6500,2.00400801603206e-05);
\draw[draw=none,fill=darkorange25512714,fill opacity=0.7] (axis cs:6500,0) rectangle (axis cs:6600,3.0060120240481e-05);
\draw[draw=none,fill=darkorange25512714,fill opacity=0.7] (axis cs:6600,0) rectangle (axis cs:6700,8.01603206412826e-05);
\draw[draw=none,fill=darkorange25512714,fill opacity=0.7] (axis cs:6700,0) rectangle (axis cs:6800,0.000220440881763527);
\draw[draw=none,fill=darkorange25512714,fill opacity=0.7] (axis cs:6800,0) rectangle (axis cs:6900,0.00032064128256513);
\draw[draw=none,fill=darkorange25512714,fill opacity=0.7] (axis cs:6900,0) rectangle (axis cs:7000,0.00031062124248497);
\draw[draw=none,fill=darkorange25512714,fill opacity=0.7] (axis cs:7000,0) rectangle (axis cs:7100,0.000160320641282565);
\draw[draw=none,fill=darkorange25512714,fill opacity=0.7] (axis cs:7100,0) rectangle (axis cs:7200,0.000100200400801603);
\draw[draw=none,fill=darkorange25512714,fill opacity=0.7] (axis cs:7200,0) rectangle (axis cs:7300,3.0060120240481e-05);
\draw[draw=none,fill=darkorange25512714,fill opacity=0.7] (axis cs:7300,0) rectangle (axis cs:7400,2.00400801603206e-05);
\draw[draw=none,fill=darkorange25512714,fill opacity=0.7] (axis cs:7400,0) rectangle (axis cs:7500,1.00200400801603e-05);
\draw[draw=none,fill=darkorange25512714,fill opacity=0.7] (axis cs:7500,0) rectangle (axis cs:7600,3.0060120240481e-05);
\draw[draw=none,fill=darkorange25512714,fill opacity=0.7] (axis cs:7600,0) rectangle (axis cs:7700,1.00200400801603e-05);
\draw[draw=none,fill=darkorange25512714,fill opacity=0.7] (axis cs:7700,0) rectangle (axis cs:7800,4.00801603206413e-05);
\draw[draw=none,fill=darkorange25512714,fill opacity=0.7] (axis cs:7800,0) rectangle (axis cs:7900,7.01402805611223e-05);
\draw[draw=none,fill=darkorange25512714,fill opacity=0.7] (axis cs:7900,0) rectangle (axis cs:8000,6.01202404809619e-05);
\draw[draw=none,fill=darkorange25512714,fill opacity=0.7] (axis cs:8000,0) rectangle (axis cs:8100,7.01402805611223e-05);
\draw[draw=none,fill=darkorange25512714,fill opacity=0.7] (axis cs:8100,0) rectangle (axis cs:8200,8.01603206412826e-05);
\draw[draw=none,fill=darkorange25512714,fill opacity=0.7] (axis cs:8200,0) rectangle (axis cs:8300,0.000100200400801603);
\draw[draw=none,fill=darkorange25512714,fill opacity=0.7] (axis cs:8300,0) rectangle (axis cs:8400,0.000100200400801603);
\draw[draw=none,fill=darkorange25512714,fill opacity=0.7] (axis cs:8400,0) rectangle (axis cs:8500,0.000190380761523046);
\draw[draw=none,fill=darkorange25512714,fill opacity=0.7] (axis cs:8500,0) rectangle (axis cs:8600,0.000130260521042084);
\draw[draw=none,fill=darkorange25512714,fill opacity=0.7] (axis cs:8600,0) rectangle (axis cs:8700,0.000100200400801603);
\draw[draw=none,fill=darkorange25512714,fill opacity=0.7] (axis cs:8700,0) rectangle (axis cs:8800,8.01603206412826e-05);
\draw[draw=none,fill=darkorange25512714,fill opacity=0.7] (axis cs:8800,0) rectangle (axis cs:8900,5.01002004008016e-05);
\draw[draw=none,fill=darkorange25512714,fill opacity=0.7] (axis cs:8900,0) rectangle (axis cs:9000,0.000110220440881764);
\draw[draw=none,fill=darkorange25512714,fill opacity=0.7] (axis cs:9000,0) rectangle (axis cs:9100,6.01202404809619e-05);
\draw[draw=none,fill=darkorange25512714,fill opacity=0.7] (axis cs:9100,0) rectangle (axis cs:9200,5.01002004008016e-05);
\draw[draw=none,fill=darkorange25512714,fill opacity=0.7] (axis cs:9200,0) rectangle (axis cs:9300,3.0060120240481e-05);
\draw[draw=none,fill=darkorange25512714,fill opacity=0.7] (axis cs:9300,0) rectangle (axis cs:9400,2.00400801603206e-05);
\draw[draw=none,fill=darkorange25512714,fill opacity=0.7] (axis cs:9400,0) rectangle (axis cs:9500,2.00400801603206e-05);
\draw[draw=none,fill=darkorange25512714,fill opacity=0.7] (axis cs:9500,0) rectangle (axis cs:9600,1.00200400801603e-05);
\draw[draw=none,fill=darkorange25512714,fill opacity=0.7] (axis cs:9600,0) rectangle (axis cs:9700,1.00200400801603e-05);
\draw[draw=none,fill=darkorange25512714,fill opacity=0.7] (axis cs:9700,0) rectangle (axis cs:9800,2.00400801603206e-05);
\draw[draw=none,fill=darkorange25512714,fill opacity=0.7] (axis cs:9800,0) rectangle (axis cs:9900,4.00801603206413e-05);
\draw[draw=none,fill=darkorange25512714,fill opacity=0.7] (axis cs:9900,0) rectangle (axis cs:10000,2.00400801603206e-05);
\draw[draw=none,fill=darkorange25512714,fill opacity=0.7] (axis cs:10000,0) rectangle (axis cs:10100,2.00400801603206e-05);
\draw[draw=none,fill=darkorange25512714,fill opacity=0.7] (axis cs:10100,0) rectangle (axis cs:10200,1.00200400801603e-05);
\draw[draw=none,fill=darkorange25512714,fill opacity=0.7] (axis cs:10200,0) rectangle (axis cs:10300,1.00200400801603e-05);
\draw[draw=none,fill=darkorange25512714,fill opacity=0.7] (axis cs:10300,0) rectangle (axis cs:10400,1.00200400801603e-05);
\draw[draw=none,fill=darkorange25512714,fill opacity=0.7] (axis cs:10400,0) rectangle (axis cs:10500,0);
\draw[draw=none,fill=darkorange25512714,fill opacity=0.7] (axis cs:10500,0) rectangle (axis cs:10600,0);
\draw[draw=none,fill=darkorange25512714,fill opacity=0.7] (axis cs:10600,0) rectangle (axis cs:10700,1.00200400801603e-05);
\draw[draw=none,fill=darkorange25512714,fill opacity=0.7] (axis cs:10700,0) rectangle (axis cs:10800,1.00200400801603e-05);
\draw[draw=none,fill=darkorange25512714,fill opacity=0.7] (axis cs:10800,0) rectangle (axis cs:10900,1.00200400801603e-05);
\draw[draw=none,fill=darkorange25512714,fill opacity=0.7] (axis cs:10900,0) rectangle (axis cs:11000,0);
\draw[draw=none,fill=darkorange25512714,fill opacity=0.7] (axis cs:11000,0) rectangle (axis cs:11100,1.00200400801603e-05);
\draw[draw=none,fill=darkorange25512714,fill opacity=0.7] (axis cs:11100,0) rectangle (axis cs:11200,1.00200400801603e-05);
\draw[draw=none,fill=darkorange25512714,fill opacity=0.7] (axis cs:11200,0) rectangle (axis cs:11300,1.00200400801603e-05);
\draw[draw=none,fill=darkorange25512714,fill opacity=0.7] (axis cs:11300,0) rectangle (axis cs:11400,1.00200400801603e-05);
\draw[draw=none,fill=darkorange25512714,fill opacity=0.7] (axis cs:11400,0) rectangle (axis cs:11500,1.00200400801603e-05);
\draw[draw=none,fill=darkorange25512714,fill opacity=0.7] (axis cs:11500,0) rectangle (axis cs:11600,0);
\draw[draw=none,fill=darkorange25512714,fill opacity=0.7] (axis cs:11600,0) rectangle (axis cs:11700,1.00200400801603e-05);
\draw[draw=none,fill=darkorange25512714,fill opacity=0.7] (axis cs:11700,0) rectangle (axis cs:11800,0);
\draw[draw=none,fill=darkorange25512714,fill opacity=0.7] (axis cs:11800,0) rectangle (axis cs:11900,1.00200400801603e-05);
\draw[draw=none,fill=darkorange25512714,fill opacity=0.7] (axis cs:11900,0) rectangle (axis cs:12000,0);
\draw[draw=none,fill=darkorange25512714,fill opacity=0.7] (axis cs:12000,0) rectangle (axis cs:12100,0);
\draw[draw=none,fill=darkorange25512714,fill opacity=0.7] (axis cs:12100,0) rectangle (axis cs:12200,1.00200400801603e-05);
\draw[draw=none,fill=darkorange25512714,fill opacity=0.7] (axis cs:12200,0) rectangle (axis cs:12300,1.00200400801603e-05);
\draw[draw=none,fill=darkorange25512714,fill opacity=0.7] (axis cs:12300,0) rectangle (axis cs:12400,0);
\draw[draw=none,fill=darkorange25512714,fill opacity=0.7] (axis cs:12400,0) rectangle (axis cs:12500,1.00200400801603e-05);
\draw[draw=none,fill=darkorange25512714,fill opacity=0.7] (axis cs:12500,0) rectangle (axis cs:12600,1.00200400801603e-05);
\draw[draw=none,fill=darkorange25512714,fill opacity=0.7] (axis cs:12600,0) rectangle (axis cs:12700,0);
\draw[draw=none,fill=darkorange25512714,fill opacity=0.7] (axis cs:12700,0) rectangle (axis cs:12800,0);
\draw[draw=none,fill=darkorange25512714,fill opacity=0.7] (axis cs:12800,0) rectangle (axis cs:12900,0);
\draw[draw=none,fill=darkorange25512714,fill opacity=0.7] (axis cs:12900,0) rectangle (axis cs:13000,0);
\draw[draw=none,fill=darkorange25512714,fill opacity=0.7] (axis cs:13000,0) rectangle (axis cs:13100,0);
\draw[draw=none,fill=darkorange25512714,fill opacity=0.7] (axis cs:13100,0) rectangle (axis cs:13200,0);
\draw[draw=none,fill=darkorange25512714,fill opacity=0.7] (axis cs:13200,0) rectangle (axis cs:13300,1.00200400801603e-05);
\draw[draw=none,fill=darkorange25512714,fill opacity=0.7] (axis cs:13300,0) rectangle (axis cs:13400,0);
\draw[draw=none,fill=darkorange25512714,fill opacity=0.7] (axis cs:13400,0) rectangle (axis cs:13500,0);
\draw[draw=none,fill=darkorange25512714,fill opacity=0.7] (axis cs:13500,0) rectangle (axis cs:13600,0);
\draw[draw=none,fill=darkorange25512714,fill opacity=0.7] (axis cs:13600,0) rectangle (axis cs:13700,0);
\draw[draw=none,fill=darkorange25512714,fill opacity=0.7] (axis cs:13700,0) rectangle (axis cs:13800,0);
\draw[draw=none,fill=darkorange25512714,fill opacity=0.7] (axis cs:13800,0) rectangle (axis cs:13900,0);
\draw[draw=none,fill=darkorange25512714,fill opacity=0.7] (axis cs:13900,0) rectangle (axis cs:14000,0);
\draw[draw=none,fill=darkorange25512714,fill opacity=0.7] (axis cs:14000,0) rectangle (axis cs:14100,0);
\draw[draw=none,fill=darkorange25512714,fill opacity=0.7] (axis cs:14100,0) rectangle (axis cs:14200,0);
\draw[draw=none,fill=darkorange25512714,fill opacity=0.7] (axis cs:14200,0) rectangle (axis cs:14300,1.00200400801603e-05);
\draw[draw=none,fill=darkorange25512714,fill opacity=0.7] (axis cs:14300,0) rectangle (axis cs:14400,0);
\draw[draw=none,fill=darkorange25512714,fill opacity=0.7] (axis cs:14400,0) rectangle (axis cs:14500,0);
\draw[draw=none,fill=darkorange25512714,fill opacity=0.7] (axis cs:14500,0) rectangle (axis cs:14600,0);
\draw[draw=none,fill=darkorange25512714,fill opacity=0.7] (axis cs:14600,0) rectangle (axis cs:14700,0);
\draw[draw=none,fill=darkorange25512714,fill opacity=0.7] (axis cs:14700,0) rectangle (axis cs:14800,0);
\draw[draw=none,fill=darkorange25512714,fill opacity=0.7] (axis cs:14800,0) rectangle (axis cs:14900,0);
\draw[draw=none,fill=darkorange25512714,fill opacity=0.7] (axis cs:14900,0) rectangle (axis cs:15000,0);
\draw[draw=none,fill=darkorange25512714,fill opacity=0.7] (axis cs:15000,0) rectangle (axis cs:15100,0);
\draw[draw=none,fill=darkorange25512714,fill opacity=0.7] (axis cs:15100,0) rectangle (axis cs:15200,0);
\draw[draw=none,fill=darkorange25512714,fill opacity=0.7] (axis cs:15200,0) rectangle (axis cs:15300,0);
\draw[draw=none,fill=darkorange25512714,fill opacity=0.7] (axis cs:15300,0) rectangle (axis cs:15400,0);
\draw[draw=none,fill=darkorange25512714,fill opacity=0.7] (axis cs:15400,0) rectangle (axis cs:15500,0);
\draw[draw=none,fill=darkorange25512714,fill opacity=0.7] (axis cs:15500,0) rectangle (axis cs:15600,0);
\draw[draw=none,fill=darkorange25512714,fill opacity=0.7] (axis cs:15600,0) rectangle (axis cs:15700,0);
\draw[draw=none,fill=darkorange25512714,fill opacity=0.7] (axis cs:15700,0) rectangle (axis cs:15800,0);
\draw[draw=none,fill=darkorange25512714,fill opacity=0.7] (axis cs:15800,0) rectangle (axis cs:15900,0);
\draw[draw=none,fill=darkorange25512714,fill opacity=0.7] (axis cs:15900,0) rectangle (axis cs:16000,0);
\draw[draw=none,fill=darkorange25512714,fill opacity=0.7] (axis cs:16000,0) rectangle (axis cs:16100,0);
\draw[draw=none,fill=darkorange25512714,fill opacity=0.7] (axis cs:16100,0) rectangle (axis cs:16200,0);
\draw[draw=none,fill=darkorange25512714,fill opacity=0.7] (axis cs:16200,0) rectangle (axis cs:16300,0);
\draw[draw=none,fill=darkorange25512714,fill opacity=0.7] (axis cs:16300,0) rectangle (axis cs:16400,0);
\draw[draw=none,fill=darkorange25512714,fill opacity=0.7] (axis cs:16400,0) rectangle (axis cs:16500,0);
\path [draw=limegreen, thick]
(axis cs:6770.49528503418,0)
--(axis cs:6770.49528503418,0.002);

\nextgroupplot[
tick pos=left,
xmin=-0.5, xmax=223.5,
y dir=reverse,
ymin=-0.5, ymax=223.5,
axis equal,
scale only axis,
height=1.136cm,
width=1.136cm,
% hide axis,
xticklabels={,,},
yticklabels={,,},
xlabel near ticks,
ylabel near ticks,
]
\addplot graphics [includegraphics cmd=\pgfimage,xmin=-0.5, xmax=223.5, ymin=223.5, ymax=-0.5] {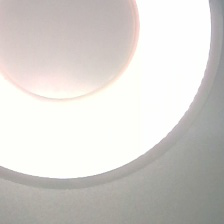};

\nextgroupplot[
tick align=outside,
tick pos=left,
x grid style={darkgray176},
xmin=0, xmax=16500,
xtick style={color=black},
xtick={0,2000,4000,6000,8000,10000,12000,14000,16000,18000},
xticklabels={$0\,\mathrm{m}$,$2\,\mathrm{m}$,$4\,\mathrm{m}$,$6\,\mathrm{m}$,$8\,\mathrm{m}$,$10\,\mathrm{m}$,$12\,\mathrm{m}$,$14\,\mathrm{m}$,$16\,\mathrm{m}$,},
y grid style={darkgray176},
ymin=0, ymax=1250,
ytick style={color=black},
ytick={0,1000},
yticklabels={$0\,\mathrm{m}$,$1\,\mathrm{m}$,},
axis equal,
scale only axis,
width=15cm,
height=1.136cm,
xtick scale label code/.code={},
ytick scale label code/.code={},
]
\addplot graphics [includegraphics cmd=\pgfimage,xmin=-0.5, xmax=16499.5, ymin=1249.5, ymax=-0.5] {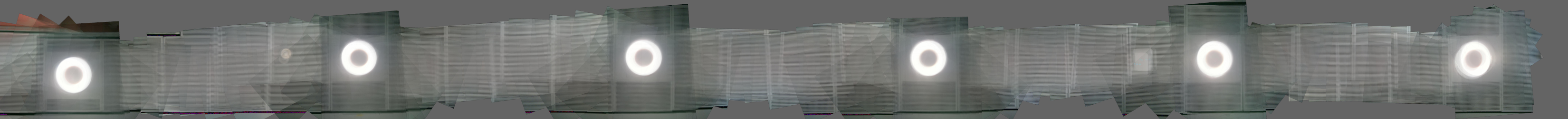};
\addplot graphics [includegraphics cmd=\pgfimage,xmin=0, xmax=16500, ymin=0, ymax=1250] {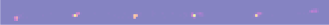};
\draw[draw=magenta,thick] (axis cs:4045.08673095703,459.515932202339) circle (150);
\draw[draw=limegreen,thick] (axis cs:6770.49528503418,517.717304825783) circle (150);
% \addplot [semithick, red]
% table {%
% 6972.60826748178 676.40192549409
% 6977.91817329063 396.144739789646
% 6693.4232771803 397.348919575285
% 6694.90305296082 674.976530031621
% 6972.60826748178 676.40192549409
% };

\nextgroupplot[
legend cell align={left},
legend style={fill opacity=0.8, draw opacity=1, text opacity=1, draw=lightgray204},
tick align=outside,
tick pos=left,
xmin=0, xmax=0.01,
y grid style={darkgray176},
ymin=0, ymax=1250,
ytick style={color=black},
scale only axis,
height=1.136cm,
width=1.136cm,
xticklabels={},
yticklabels={},
ticks=none,
xtick scale label code/.code={},
xlabel near ticks,
ylabel near ticks,
hide axis,
]
\draw[draw=none,fill=steelblue31119180,fill opacity=0.9] (axis cs:0,7.105427357601e-15) rectangle (axis cs:0,104.166666666667);
\draw[draw=none,fill=steelblue31119180,fill opacity=0.9] (axis cs:0,104.166666666667) rectangle (axis cs:0,208.333333333333);
\draw[draw=none,fill=steelblue31119180,fill opacity=0.9] (axis cs:0,208.333333333333) rectangle (axis cs:0.0004992,312.5);
\draw[draw=none,fill=steelblue31119180,fill opacity=0.9] (axis cs:0,312.5) rectangle (axis cs:0.0016032,416.666666666667);
\draw[draw=none,fill=steelblue31119180,fill opacity=0.9] (axis cs:0,416.666666666667) rectangle (axis cs:0.0049248,520.833333333333);
\draw[draw=none,fill=steelblue31119180,fill opacity=0.9] (axis cs:0,520.833333333333) rectangle (axis cs:0.0017472,625);
\draw[draw=none,fill=steelblue31119180,fill opacity=0.9] (axis cs:0,625) rectangle (axis cs:0.000672,729.166666666667);
\draw[draw=none,fill=steelblue31119180,fill opacity=0.9] (axis cs:0,729.166666666667) rectangle (axis cs:0.0001536,833.333333333333);
\draw[draw=none,fill=steelblue31119180,fill opacity=0.9] (axis cs:0,833.333333333333) rectangle (axis cs:0,937.5);
\draw[draw=none,fill=steelblue31119180,fill opacity=0.9] (axis cs:0,937.5) rectangle (axis cs:0,1041.66666666667);
\draw[draw=none,fill=steelblue31119180,fill opacity=0.9] (axis cs:0,1041.66666666667) rectangle (axis cs:0,1145.83333333333);
\draw[draw=none,fill=steelblue31119180,fill opacity=0.9] (axis cs:0,1145.83333333333) rectangle (axis cs:0,1250);
\path [draw=magenta, thick]
(axis cs:0,459.515932202339)
--(axis cs:0.01,459.515932202339);

\draw[draw=none,fill=darkorange25512714,fill opacity=0.7] (axis cs:0,7.105427357601e-15) rectangle (axis cs:0,104.166666666667);
\draw[draw=none,fill=darkorange25512714,fill opacity=0.7] (axis cs:0,104.166666666667) rectangle (axis cs:0,208.333333333333);
\draw[draw=none,fill=darkorange25512714,fill opacity=0.7] (axis cs:0,208.333333333333) rectangle (axis cs:0,312.5);
\draw[draw=none,fill=darkorange25512714,fill opacity=0.7] (axis cs:0,312.5) rectangle (axis cs:0.0016128,416.666666666667);
\draw[draw=none,fill=darkorange25512714,fill opacity=0.7] (axis cs:0,416.666666666667) rectangle (axis cs:0.0079872,520.833333333333);
\draw[draw=none,fill=darkorange25512714,fill opacity=0.7] (axis cs:0,520.833333333333) rectangle (axis cs:0,625);
\draw[draw=none,fill=darkorange25512714,fill opacity=0.7] (axis cs:0,625) rectangle (axis cs:0,729.166666666667);
\draw[draw=none,fill=darkorange25512714,fill opacity=0.7] (axis cs:0,729.166666666667) rectangle (axis cs:0,833.333333333333);
\draw[draw=none,fill=darkorange25512714,fill opacity=0.7] (axis cs:0,833.333333333333) rectangle (axis cs:0,937.5);
\draw[draw=none,fill=darkorange25512714,fill opacity=0.7] (axis cs:0,937.5) rectangle (axis cs:0,1041.66666666667);
\draw[draw=none,fill=darkorange25512714,fill opacity=0.7] (axis cs:0,1041.66666666667) rectangle (axis cs:0,1145.83333333333);
\draw[draw=none,fill=darkorange25512714,fill opacity=0.7] (axis cs:0,1145.83333333333) rectangle (axis cs:0,1250);
\path [draw=limegreen, thick]
(axis cs:0,517.717304825783)
--(axis cs:0.01,517.717304825783);
\end{groupplot}%
\end{tikzpicture}%

%% file: figures/synthetic_ablation/ablation_round_1000.tex
\footnotesize

% This file was created with tikzplotlib v0.10.1.
\begin{tikzpicture}

\definecolor{darkgray176}{RGB}{176,176,176}

\definecolor{materialviolet}{RGB}{98, 0, 238}

\begin{axis}[
width=3.5cm,
height=2.6cm,
scale only axis=true,
xmajorgrids,
ymajorgrids,
tick align=outside,
tick pos=left,
x grid style={darkgray176},
xmin=0.5, xmax=21.5,
xtick style={color=black},
xtick={1,5,9,13,17,21},
xticklabels={0.01,0.2,0.4,0.6,0.8,1.0},
y grid style={darkgray176},
ymin=-0.05, ymax=1.05,
ytick style={color=black},
xlabel={$\alpha$},
ticklabel style = {font=\scriptsize},
yticklabel=$\smash{\pgfmathprintnumber{\tick}}$,
]
\addplot [draw=black, fill=materialteal!30]
table {%
0.75 0
1.25 0
1.25 0
0.75 0
0.75 0
};
\addplot [draw=black, fill=materialteal!30]
table {%
1 0
1 0
};
\addplot [draw=black, fill=materialteal!30]
table {%
1 0
1 0
};
\addplot [draw=black, fill=materialteal!30]
table {%
0.875 0
1.125 0
};
\addplot [draw=black, fill=materialteal!30]
table {%
0.875 0
1.125 0
};
\addplot [draw=black, fill=materialteal!30]
table {%
1.75 0.0180000006221235
2.25 0.0180000006221235
2.25 0.0940000023692846
1.75 0.0940000023692846
1.75 0.0180000006221235
};
\addplot [draw=black, fill=materialteal!30]
table {%
2 0.0180000006221235
2 0
};
\addplot [draw=black, fill=materialteal!30]
table {%
2 0.0940000023692846
2 0.208000004291534
};
\addplot [draw=black, fill=materialteal!30]
table {%
1.875 0
2.125 0
};
\addplot [draw=black, fill=materialteal!30]
table {%
1.875 0.208000004291534
2.125 0.208000004291534
};
\addplot [draw=black, fill=materialteal!30]
table {%
2.75 0.392000027000904
3.25 0.392000027000904
3.25 0.676000013947487
2.75 0.676000013947487
2.75 0.392000027000904
};
\addplot [draw=black, fill=materialteal!30]
table {%
3 0.392000027000904
3 0.312000006437302
};
\addplot [draw=black, fill=materialteal!30]
table {%
3 0.676000013947487
3 0.76800000667572
};
\addplot [draw=black, fill=materialteal!30]
table {%
2.875 0.312000006437302
3.125 0.312000006437302
};
\addplot [draw=black, fill=materialteal!30]
table {%
2.875 0.76800000667572
3.125 0.76800000667572
};
\addplot [draw=black, fill=materialteal!30]
table {%
3.75 0.464000023901463
4.25 0.464000023901463
4.25 0.776000022888184
3.75 0.776000022888184
3.75 0.464000023901463
};
\addplot [draw=black, fill=materialteal!30]
table {%
4 0.464000023901463
4 0.248000010848045
};
\addplot [draw=black, fill=materialteal!30]
table {%
4 0.776000022888184
4 0.888000071048737
};
\addplot [draw=black, fill=materialteal!30]
table {%
3.875 0.248000010848045
4.125 0.248000010848045
};
\addplot [draw=black, fill=materialteal!30]
table {%
3.875 0.888000071048737
4.125 0.888000071048737
};
\addplot [draw=black, fill=materialteal!30]
table {%
4.75 0.338000021874905
5.25 0.338000021874905
5.25 0.708000034093857
4.75 0.708000034093857
4.75 0.338000021874905
};
\addplot [draw=black, fill=materialteal!30]
table {%
5 0.338000021874905
5 0.224000006914139
};
\addplot [draw=black, fill=materialteal!30]
table {%
5 0.708000034093857
5 0.840000033378601
};
\addplot [draw=black, fill=materialteal!30]
table {%
4.875 0.224000006914139
5.125 0.224000006914139
};
\addplot [draw=black, fill=materialteal!30]
table {%
4.875 0.840000033378601
5.125 0.840000033378601
};
\addplot [draw=black, fill=materialteal!30]
table {%
5.75 0.202000014483929
6.25 0.202000014483929
6.25 0.554000005125999
5.75 0.554000005125999
5.75 0.202000014483929
};
\addplot [draw=black, fill=materialteal!30]
table {%
6 0.202000014483929
6 0.112000003457069
};
\addplot [draw=black, fill=materialteal!30]
table {%
6 0.554000005125999
6 0.784000039100647
};
\addplot [draw=black, fill=materialteal!30]
table {%
5.875 0.112000003457069
6.125 0.112000003457069
};
\addplot [draw=black, fill=materialteal!30]
table {%
5.875 0.784000039100647
6.125 0.784000039100647
};
\addplot [draw=black, fill=materialteal!30]
table {%
6.75 0.242000009864569
7.25 0.242000009864569
7.25 0.458000026643276
6.75 0.458000026643276
6.75 0.242000009864569
};
\addplot [draw=black, fill=materialteal!30]
table {%
7 0.242000009864569
7 0.00800000037997961
};
\addplot [draw=black, fill=materialteal!30]
table {%
7 0.458000026643276
7 0.688000023365021
};
\addplot [draw=black, fill=materialteal!30]
table {%
6.875 0.00800000037997961
7.125 0.00800000037997961
};
\addplot [draw=black, fill=materialteal!30]
table {%
6.875 0.688000023365021
7.125 0.688000023365021
};
\addplot [draw=black, fill=materialteal!30]
table {%
7.75 0
8.25 0
8.25 0.0880000023171306
7.75 0.0880000023171306
7.75 0
};
\addplot [draw=black, fill=materialteal!30]
table {%
8 0
8 0
};
\addplot [draw=black, fill=materialteal!30]
table {%
8 0.0880000023171306
8 0.232000008225441
};
\addplot [draw=black, fill=materialteal!30]
table {%
7.875 0
8.125 0
};
\addplot [draw=black, fill=materialteal!30]
table {%
7.875 0.232000008225441
8.125 0.232000008225441
};
\addplot [draw=black, fill=materialteal!30]
table {%
8.75 0
9.25 0
9.25 0
8.75 0
8.75 0
};
\addplot [draw=black, fill=materialteal!30]
table {%
9 0
9 0
};
\addplot [draw=black, fill=materialteal!30]
table {%
9 0
9 0.00800000037997961
};
\addplot [draw=black, fill=materialteal!30]
table {%
8.875 0
9.125 0
};
\addplot [draw=black, fill=materialteal!30]
table {%
8.875 0.00800000037997961
9.125 0.00800000037997961
};
\addplot [draw=black, fill=materialteal!30]
table {%
9.75 0
10.25 0
10.25 0.0480000022798777
9.75 0.0480000022798777
9.75 0
};
\addplot [draw=black, fill=materialteal!30]
table {%
10 0
10 0
};
\addplot [draw=black, fill=materialteal!30]
table {%
10 0.0480000022798777
10 0.240000009536743
};
\addplot [draw=black, fill=materialteal!30]
table {%
9.875 0
10.125 0
};
\addplot [draw=black, fill=materialteal!30]
table {%
9.875 0.240000009536743
10.125 0.240000009536743
};
\addplot [draw=black, fill=materialteal!30]
table {%
10.75 0
11.25 0
11.25 0.0160000007599592
10.75 0.0160000007599592
10.75 0
};
\addplot [draw=black, fill=materialteal!30]
table {%
11 0
11 0
};
\addplot [draw=black, fill=materialteal!30]
table {%
11 0.0160000007599592
11 0.112000003457069
};
\addplot [draw=black, fill=materialteal!30]
table {%
10.875 0
11.125 0
};
\addplot [draw=black, fill=materialteal!30]
table {%
10.875 0.112000003457069
11.125 0.112000003457069
};
\addplot [draw=black, fill=materialteal!30]
table {%
11.75 0.00400000018998981
12.25 0.00400000018998981
12.25 0.0660000033676624
11.75 0.0660000033676624
11.75 0.00400000018998981
};
\addplot [draw=black, fill=materialteal!30]
table {%
12 0.00400000018998981
12 0
};
\addplot [draw=black, fill=materialteal!30]
table {%
12 0.0660000033676624
12 0.15200001001358
};
\addplot [draw=black, fill=materialteal!30]
table {%
11.875 0
12.125 0
};
\addplot [draw=black, fill=materialteal!30]
table {%
11.875 0.15200001001358
12.125 0.15200001001358
};
\addplot [draw=black, fill=materialteal!30]
table {%
12.75 0
13.25 0
13.25 0
12.75 0
12.75 0
};
\addplot [draw=black, fill=materialteal!30]
table {%
13 0
13 0
};
\addplot [draw=black, fill=materialteal!30]
table {%
13 0
13 0.0320000015199184
};
\addplot [draw=black, fill=materialteal!30]
table {%
12.875 0
13.125 0
};
\addplot [draw=black, fill=materialteal!30]
table {%
12.875 0.0320000015199184
13.125 0.0320000015199184
};
\addplot [draw=black, fill=materialteal!30]
table {%
13.75 0
14.25 0
14.25 0
13.75 0
13.75 0
};
\addplot [draw=black, fill=materialteal!30]
table {%
14 0
14 0
};
\addplot [draw=black, fill=materialteal!30]
table {%
14 0
14 0
};
\addplot [draw=black, fill=materialteal!30]
table {%
13.875 0
14.125 0
};
\addplot [draw=black, fill=materialteal!30]
table {%
13.875 0
14.125 0
};
\addplot [draw=black, fill=materialteal!30]
table {%
14.75 0
15.25 0
15.25 0
14.75 0
14.75 0
};
\addplot [draw=black, fill=materialteal!30]
table {%
15 0
15 0
};
\addplot [draw=black, fill=materialteal!30]
table {%
15 0
15 0.0480000004172325
};
\addplot [draw=black, fill=materialteal!30]
table {%
14.875 0
15.125 0
};
\addplot [draw=black, fill=materialteal!30]
table {%
14.875 0.0480000004172325
15.125 0.0480000004172325
};
\addplot [draw=black, fill=materialteal!30]
table {%
15.75 0
16.25 0
16.25 0
15.75 0
15.75 0
};
\addplot [draw=black, fill=materialteal!30]
table {%
16 0
16 0
};
\addplot [draw=black, fill=materialteal!30]
table {%
16 0
16 0
};
\addplot [draw=black, fill=materialteal!30]
table {%
15.875 0
16.125 0
};
\addplot [draw=black, fill=materialteal!30]
table {%
15.875 0
16.125 0
};
\addplot [draw=black, fill=materialteal!30]
table {%
16.75 0
17.25 0
17.25 0
16.75 0
16.75 0
};
\addplot [draw=black, fill=materialteal!30]
table {%
17 0
17 0
};
\addplot [draw=black, fill=materialteal!30]
table {%
17 0
17 0
};
\addplot [draw=black, fill=materialteal!30]
table {%
16.875 0
17.125 0
};
\addplot [draw=black, fill=materialteal!30]
table {%
16.875 0
17.125 0
};
\addplot [draw=black, fill=materialteal!30]
table {%
17.75 0
18.25 0
18.25 0
17.75 0
17.75 0
};
\addplot [draw=black, fill=materialteal!30]
table {%
18 0
18 0
};
\addplot [draw=black, fill=materialteal!30]
table {%
18 0
18 0
};
\addplot [draw=black, fill=materialteal!30]
table {%
17.875 0
18.125 0
};
\addplot [draw=black, fill=materialteal!30]
table {%
17.875 0
18.125 0
};
\addplot [draw=black, fill=materialteal!30]
table {%
18.75 0
19.25 0
19.25 0
18.75 0
18.75 0
};
\addplot [draw=black, fill=materialteal!30]
table {%
19 0
19 0
};
\addplot [draw=black, fill=materialteal!30]
table {%
19 0
19 0
};
\addplot [draw=black, fill=materialteal!30]
table {%
18.875 0
19.125 0
};
\addplot [draw=black, fill=materialteal!30]
table {%
18.875 0
19.125 0
};
\addplot [draw=black, fill=materialteal!30]
table {%
19.75 0
20.25 0
20.25 0
19.75 0
19.75 0
};
\addplot [draw=black, fill=materialteal!30]
table {%
20 0
20 0
};
\addplot [draw=black, fill=materialteal!30]
table {%
20 0
20 0
};
\addplot [draw=black, fill=materialteal!30]
table {%
19.875 0
20.125 0
};
\addplot [draw=black, fill=materialteal!30]
table {%
19.875 0
20.125 0
};
\addplot [draw=black, fill=materialteal!30]
table {%
20.75 0
21.25 0
21.25 0
20.75 0
20.75 0
};
\addplot [draw=black, fill=materialteal!30]
table {%
21 0
21 0
};
\addplot [draw=black, fill=materialteal!30]
table {%
21 0
21 0.00800000037997961
};
\addplot [draw=black, fill=materialteal!30]
table {%
20.875 0
21.125 0
};
\addplot [draw=black, fill=materialteal!30]
table {%
20.875 0.00800000037997961
21.125 0.00800000037997961
};
\addplot [materialviolet, thick]
table {%
0.75 0
1.25 0
};
\addplot [materialviolet, thick]
table {%
1.75 0.0480000022798777
2.25 0.0480000022798777
};
\addplot [materialviolet, thick]
table {%
2.75 0.504000023007393
3.25 0.504000023007393
};
\addplot [materialviolet, thick]
table {%
3.75 0.692000031471252
4.25 0.692000031471252
};
\addplot [materialviolet, thick]
table {%
4.75 0.468000009655952
5.25 0.468000009655952
};
\addplot [materialviolet, thick]
table {%
5.75 0.452000021934509
6.25 0.452000021934509
};
\addplot [materialviolet, thick]
table {%
6.75 0.372000023722649
7.25 0.372000023722649
};
\addplot [materialviolet, thick]
table {%
7.75 0
8.25 0
};
\addplot [materialviolet, thick]
table {%
8.75 0
9.25 0
};
\addplot [materialviolet, thick]
table {%
9.75 0
10.25 0
};
\addplot [materialviolet, thick]
table {%
10.75 0.00800000037997961
11.25 0.00800000037997961
};
\addplot [materialviolet, thick]
table {%
11.75 0.0400000028312206
12.25 0.0400000028312206
};
\addplot [materialviolet, thick]
table {%
12.75 0
13.25 0
};
\addplot [materialviolet, thick]
table {%
13.75 0
14.25 0
};
\addplot [materialviolet, thick]
table {%
14.75 0
15.25 0
};
\addplot [materialviolet, thick]
table {%
15.75 0
16.25 0
};
\addplot [materialviolet, thick]
table {%
16.75 0
17.25 0
};
\addplot [materialviolet, thick]
table {%
17.75 0
18.25 0
};
\addplot [materialviolet, thick]
table {%
18.75 0
19.25 0
};
\addplot [materialviolet, thick]
table {%
19.75 0
20.25 0
};
\addplot [materialviolet, thick]
table {%
20.75 0
21.25 0
};
\end{axis}

\end{tikzpicture}

%% file: figures/synthetic_ablation/ablation_long_1000.tex
\footnotesize

% This file was created with tikzplotlib v0.10.1.
\begin{tikzpicture}

\definecolor{darkgray176}{RGB}{176,176,176}

\definecolor{materialviolet}{RGB}{98, 0, 238}

\begin{axis}[
width=3.5cm,
height=2.6cm,
scale only axis=true,
xmajorgrids,
ymajorgrids,
tick align=outside,
tick pos=left,
x grid style={darkgray176},
xmin=0.5, xmax=21.5,
xtick style={color=black},
xtick={1,5,9,13,17,21},
xticklabels={0.01,0.2,0.4,0.6,0.8,1.0},
y grid style={darkgray176},
ymin=-0.05, ymax=1.05,
ytick style={color=black},
yticklabels={,,},
ytick style={draw=none},
xlabel={$\alpha$},
ticklabel style = {font=\scriptsize},
]
\addplot [draw=black, fill=materialteal!30]
table {%
0.75 0
1.25 0
1.25 0
0.75 0
0.75 0
};
\addplot [draw=black, fill=materialteal!30]
table {%
1 0
1 0
};
\addplot [draw=black, fill=materialteal!30]
table {%
1 0
1 0
};
\addplot [draw=black, fill=materialteal!30]
table {%
0.875 0
1.125 0
};
\addplot [draw=black, fill=materialteal!30]
table {%
0.875 0
1.125 0
};
\addplot [draw=black, fill=materialteal!30]
table {%
1.75 0.326000012457371
2.25 0.326000012457371
2.25 0.558000013232231
1.75 0.558000013232231
1.75 0.326000012457371
};
\addplot [draw=black, fill=materialteal!30]
table {%
2 0.326000012457371
2 0.272000014781952
};
\addplot [draw=black, fill=materialteal!30]
table {%
2 0.558000013232231
2 0.616000056266785
};
\addplot [draw=black, fill=materialteal!30]
table {%
1.875 0.272000014781952
2.125 0.272000014781952
};
\addplot [draw=black, fill=materialteal!30]
table {%
1.875 0.616000056266785
2.125 0.616000056266785
};
\addplot [draw=black, fill=materialteal!30]
table {%
2.75 0.936000034213066
3.25 0.936000034213066
3.25 0.990000039339066
2.75 0.990000039339066
2.75 0.936000034213066
};
\addplot [draw=black, fill=materialteal!30]
table {%
3 0.936000034213066
3 0.864000022411346
};
\addplot [draw=black, fill=materialteal!30]
table {%
3 0.990000039339066
3 1
};
\addplot [draw=black, fill=materialteal!30]
table {%
2.875 0.864000022411346
3.125 0.864000022411346
};
\addplot [draw=black, fill=materialteal!30]
table {%
2.875 1
3.125 1
};
\addplot [draw=black, fill=materialteal!30]
table {%
3.75 0.960000038146973
4.25 0.960000038146973
4.25 0.990000039339066
3.75 0.990000039339066
3.75 0.960000038146973
};
\addplot [draw=black, fill=materialteal!30]
table {%
4 0.960000038146973
4 0.936000049114227
};
\addplot [draw=black, fill=materialteal!30]
table {%
4 0.990000039339066
4 1
};
\addplot [draw=black, fill=materialteal!30]
table {%
3.875 0.936000049114227
4.125 0.936000049114227
};
\addplot [draw=black, fill=materialteal!30]
table {%
3.875 1
4.125 1
};
\addplot [draw=black, fill=materialteal!30]
table {%
4.75 0.938000053167343
5.25 0.938000053167343
5.25 1
4.75 1
4.75 0.938000053167343
};
\addplot [draw=black, fill=materialteal!30]
table {%
5 0.938000053167343
5 0.888000071048737
};
\addplot [draw=black, fill=materialteal!30]
table {%
5 1
5 1
};
\addplot [draw=black, fill=materialteal!30]
table {%
4.875 0.888000071048737
5.125 0.888000071048737
};
\addplot [draw=black, fill=materialteal!30]
table {%
4.875 1
5.125 1
};
\addplot [draw=black, fill=materialteal!30]
table {%
5.75 0.860000044107437
6.25 0.860000044107437
6.25 0.976000070571899
5.75 0.976000070571899
5.75 0.860000044107437
};
\addplot [draw=black, fill=materialteal!30]
table {%
6 0.860000044107437
6 0.79200005531311
};
\addplot [draw=black, fill=materialteal!30]
table {%
6 0.976000070571899
6 0.984000027179718
};
\addplot [draw=black, fill=materialteal!30]
table {%
5.875 0.79200005531311
6.125 0.79200005531311
};
\addplot [draw=black, fill=materialteal!30]
table {%
5.875 0.984000027179718
6.125 0.984000027179718
};
\addplot [draw=black, fill=materialteal!30]
table {%
6.75 0.922000020742416
7.25 0.922000020742416
7.25 0.976000070571899
6.75 0.976000070571899
6.75 0.922000020742416
};
\addplot [draw=black, fill=materialteal!30]
table {%
7 0.922000020742416
7 0.656000018119812
};
\addplot [draw=black, fill=materialteal!30]
table {%
7 0.976000070571899
7 0.992000043392181
};
\addplot [draw=black, fill=materialteal!30]
table {%
6.875 0.656000018119812
7.125 0.656000018119812
};
\addplot [draw=black, fill=materialteal!30]
table {%
6.875 0.992000043392181
7.125 0.992000043392181
};
\addplot [draw=black, fill=materialteal!30]
table {%
7.75 0.922000020742416
8.25 0.922000020742416
8.25 0.950000032782555
7.75 0.950000032782555
7.75 0.922000020742416
};
\addplot [draw=black, fill=materialteal!30]
table {%
8 0.922000020742416
8 0.616000056266785
};
\addplot [draw=black, fill=materialteal!30]
table {%
8 0.950000032782555
8 0.992000043392181
};
\addplot [draw=black, fill=materialteal!30]
table {%
7.875 0.616000056266785
8.125 0.616000056266785
};
\addplot [draw=black, fill=materialteal!30]
table {%
7.875 0.992000043392181
8.125 0.992000043392181
};
\addplot [draw=black, fill=materialteal!30]
table {%
8.75 0.688000038266182
9.25 0.688000038266182
9.25 0.948000028729439
8.75 0.948000028729439
8.75 0.688000038266182
};
\addplot [draw=black, fill=materialteal!30]
table {%
9 0.688000038266182
9 0.552000045776367
};
\addplot [draw=black, fill=materialteal!30]
table {%
9 0.948000028729439
9 0.992000043392181
};
\addplot [draw=black, fill=materialteal!30]
table {%
8.875 0.552000045776367
9.125 0.552000045776367
};
\addplot [draw=black, fill=materialteal!30]
table {%
8.875 0.992000043392181
9.125 0.992000043392181
};
\addplot [draw=black, fill=materialteal!30]
table {%
9.75 0.740000039339066
10.25 0.740000039339066
10.25 0.866000041365623
9.75 0.866000041365623
9.75 0.740000039339066
};
\addplot [draw=black, fill=materialteal!30]
table {%
10 0.740000039339066
10 0.648000001907349
};
\addplot [draw=black, fill=materialteal!30]
table {%
10 0.866000041365623
10 0.960000038146973
};
\addplot [draw=black, fill=materialteal!30]
table {%
9.875 0.648000001907349
10.125 0.648000001907349
};
\addplot [draw=black, fill=materialteal!30]
table {%
9.875 0.960000038146973
10.125 0.960000038146973
};
\addplot [draw=black, fill=materialteal!30]
table {%
10.75 0.650000035762787
11.25 0.650000035762787
11.25 0.856000036001205
10.75 0.856000036001205
10.75 0.650000035762787
};
\addplot [draw=black, fill=materialteal!30]
table {%
11 0.650000035762787
11 0.496000021696091
};
\addplot [draw=black, fill=materialteal!30]
table {%
11 0.856000036001205
11 0.920000016689301
};
\addplot [draw=black, fill=materialteal!30]
table {%
10.875 0.496000021696091
11.125 0.496000021696091
};
\addplot [draw=black, fill=materialteal!30]
table {%
10.875 0.920000016689301
11.125 0.920000016689301
};
\addplot [draw=black, fill=materialteal!30]
table {%
11.75 0.696000024676323
12.25 0.696000024676323
12.25 0.916000038385391
11.75 0.916000038385391
11.75 0.696000024676323
};
\addplot [draw=black, fill=materialteal!30]
table {%
12 0.696000024676323
12 0.608000040054321
};
\addplot [draw=black, fill=materialteal!30]
table {%
12 0.916000038385391
12 0.928000032901764
};
\addplot [draw=black, fill=materialteal!30]
table {%
11.875 0.608000040054321
12.125 0.608000040054321
};
\addplot [draw=black, fill=materialteal!30]
table {%
11.875 0.928000032901764
12.125 0.928000032901764
};
\addplot [draw=black, fill=materialteal!30]
table {%
12.75 0.59400001168251
13.25 0.59400001168251
13.25 0.802000030875206
12.75 0.802000030875206
12.75 0.59400001168251
};
\addplot [draw=black, fill=materialteal!30]
table {%
13 0.59400001168251
13 0.248000010848045
};
\addplot [draw=black, fill=materialteal!30]
table {%
13 0.802000030875206
13 0.840000033378601
};
\addplot [draw=black, fill=materialteal!30]
table {%
12.875 0.248000010848045
13.125 0.248000010848045
};
\addplot [draw=black, fill=materialteal!30]
table {%
12.875 0.840000033378601
13.125 0.840000033378601
};
\addplot [draw=black, fill=materialteal!30]
table {%
13.75 0.55000002682209
14.25 0.55000002682209
14.25 0.68400003015995
13.75 0.68400003015995
13.75 0.55000002682209
};
\addplot [draw=black, fill=materialteal!30]
table {%
14 0.55000002682209
14 0.176000013947487
};
\addplot [draw=black, fill=materialteal!30]
table {%
14 0.68400003015995
14 0.776000022888184
};
\addplot [draw=black, fill=materialteal!30]
table {%
13.875 0.176000013947487
14.125 0.176000013947487
};
\addplot [draw=black, fill=materialteal!30]
table {%
13.875 0.776000022888184
14.125 0.776000022888184
};
\addplot [draw=black, fill=materialteal!30]
table {%
14.75 0.242000006139278
15.25 0.242000006139278
15.25 0.520000003278255
14.75 0.520000003278255
14.75 0.242000006139278
};
\addplot [draw=black, fill=materialteal!30]
table {%
15 0.242000006139278
15 0.160000011324883
};
\addplot [draw=black, fill=materialteal!30]
table {%
15 0.520000003278255
15 0.544000029563904
};
\addplot [draw=black, fill=materialteal!30]
table {%
14.875 0.160000011324883
15.125 0.160000011324883
};
\addplot [draw=black, fill=materialteal!30]
table {%
14.875 0.544000029563904
15.125 0.544000029563904
};
\addplot [draw=black, fill=materialteal!30]
table {%
15.75 0.104000002145767
16.25 0.104000002145767
16.25 0.26800000295043
15.75 0.26800000295043
15.75 0.104000002145767
};
\addplot [draw=black, fill=materialteal!30]
table {%
16 0.104000002145767
16 0.0320000015199184
};
\addplot [draw=black, fill=materialteal!30]
table {%
16 0.26800000295043
16 0.400000005960464
};
\addplot [draw=black, fill=materialteal!30]
table {%
15.875 0.0320000015199184
16.125 0.0320000015199184
};
\addplot [draw=black, fill=materialteal!30]
table {%
15.875 0.400000005960464
16.125 0.400000005960464
};
\addplot [draw=black, fill=materialteal!30]
table {%
16.75 0
17.25 0
17.25 0.0140000006649643
16.75 0.0140000006649643
16.75 0
};
\addplot [draw=black, fill=materialteal!30]
table {%
17 0
17 0
};
\addplot [draw=black, fill=materialteal!30]
table {%
17 0.0140000006649643
17 0.0480000004172325
};
\addplot [draw=black, fill=materialteal!30]
table {%
16.875 0
17.125 0
};
\addplot [draw=black, fill=materialteal!30]
table {%
16.875 0.0480000004172325
17.125 0.0480000004172325
};
\addplot [draw=black, fill=materialteal!30]
table {%
17.75 0
18.25 0
18.25 0
17.75 0
17.75 0
};
\addplot [draw=black, fill=materialteal!30]
table {%
18 0
18 0
};
\addplot [draw=black, fill=materialteal!30]
table {%
18 0
18 0
};
\addplot [draw=black, fill=materialteal!30]
table {%
17.875 0
18.125 0
};
\addplot [draw=black, fill=materialteal!30]
table {%
17.875 0
18.125 0
};
\addplot [draw=black, fill=materialteal!30]
table {%
18.75 0
19.25 0
19.25 0
18.75 0
18.75 0
};
\addplot [draw=black, fill=materialteal!30]
table {%
19 0
19 0
};
\addplot [draw=black, fill=materialteal!30]
table {%
19 0
19 0
};
\addplot [draw=black, fill=materialteal!30]
table {%
18.875 0
19.125 0
};
\addplot [draw=black, fill=materialteal!30]
table {%
18.875 0
19.125 0
};
\addplot [draw=black, fill=materialteal!30]
table {%
19.75 0
20.25 0
20.25 0
19.75 0
19.75 0
};
\addplot [draw=black, fill=materialteal!30]
table {%
20 0
20 0
};
\addplot [draw=black, fill=materialteal!30]
table {%
20 0
20 0
};
\addplot [draw=black, fill=materialteal!30]
table {%
19.875 0
20.125 0
};
\addplot [draw=black, fill=materialteal!30]
table {%
19.875 0
20.125 0
};
\addplot [draw=black, fill=materialteal!30]
table {%
20.75 0
21.25 0
21.25 0
20.75 0
20.75 0
};
\addplot [draw=black, fill=materialteal!30]
table {%
21 0
21 0
};
\addplot [draw=black, fill=materialteal!30]
table {%
21 0
21 0
};
\addplot [draw=black, fill=materialteal!30]
table {%
20.875 0
21.125 0
};
\addplot [draw=black, fill=materialteal!30]
table {%
20.875 0
21.125 0
};
\addplot [materialviolet, thick]
table {%
0.75 0
1.25 0
};
\addplot [materialviolet, thick]
table {%
1.75 0.456000030040741
2.25 0.456000030040741
};
\addplot [materialviolet, thick]
table {%
2.75 0.980000048875809
3.25 0.980000048875809
};
\addplot [materialviolet, thick]
table {%
3.75 0.976000070571899
4.25 0.976000070571899
};
\addplot [materialviolet, thick]
table {%
4.75 0.980000048875809
5.25 0.980000048875809
};
\addplot [materialviolet, thick]
table {%
5.75 0.940000057220459
6.25 0.940000057220459
};
\addplot [materialviolet, thick]
table {%
6.75 0.944000035524368
7.25 0.944000035524368
};
\addplot [materialviolet, thick]
table {%
7.75 0.936000049114227
8.25 0.936000049114227
};
\addplot [materialviolet, thick]
table {%
8.75 0.860000044107437
9.25 0.860000044107437
};
\addplot [materialviolet, thick]
table {%
9.75 0.79600003361702
10.25 0.79600003361702
};
\addplot [materialviolet, thick]
table {%
10.75 0.744000047445297
11.25 0.744000047445297
};
\addplot [materialviolet, thick]
table {%
11.75 0.848000049591064
12.25 0.848000049591064
};
\addplot [materialviolet, thick]
table {%
12.75 0.776000022888184
13.25 0.776000022888184
};
\addplot [materialviolet, thick]
table {%
13.75 0.620000034570694
14.25 0.620000034570694
};
\addplot [materialviolet, thick]
table {%
14.75 0.480000019073486
15.25 0.480000019073486
};
\addplot [materialviolet, thick]
table {%
15.75 0.216000005602837
16.25 0.216000005602837
};
\addplot [materialviolet, thick]
table {%
16.75 0.00400000018998981
17.25 0.00400000018998981
};
\addplot [materialviolet, thick]
table {%
17.75 0
18.25 0
};
\addplot [materialviolet, thick]
table {%
18.75 0
19.25 0
};
\addplot [materialviolet, thick]
table {%
19.75 0
20.25 0
};
\addplot [materialviolet, thick]
table {%
20.75 0
21.25 0
};
\end{axis}

\end{tikzpicture}